\PassOptionsToPackage{dvipsnames}{xcolor}
\documentclass{article}
\usepackage{iclr2027_conference,times}

\usepackage[T1]{fontenc}
\usepackage[utf8]{inputenc}
\usepackage{amsmath,amssymb}
\usepackage{bm}
\usepackage{graphicx}
\usepackage{booktabs}
\usepackage{longtable}
\usepackage{placeins}
\usepackage{tikz}
\usetikzlibrary{positioning, arrows.meta, fit, calc}
\usepackage[protrusion=true,expansion=true,kerning]{microtype}
\usepackage{hyperref}
\usepackage{url}

\tikzset{>=latex}

\newcommand{\secref}[1]{Section~\ref{#1}}
\newcommand{\figref}[1]{Figure~\ref{#1}}
\newcommand{\tabref}[1]{Table~\ref{#1}}
\renewcommand{\eqref}[1]{Equation~(\ref{#1})}
\newcommand{\appref}[1]{Appendix~\ref{#1}}

\title{Improving Neural Symbolic Regression with a Diverse Prior and Post-training Alignment}

\author{
  David Otte$^{1,2}$ \thanks{Work done while at the University of Freiburg.} \quad
  Jörg K.H. Franke$^{3,4}$ \quad
  Frank Hutter$^{3,5,6}$ \quad
  \\ \\
  $^1$LMU Munich \quad
  $^2$Munich Center for Machine Learning (MCML) \quad
  $^3$University of Freiburg \quad \\
  $^4$LAION \quad
  $^5$ELLIS Institute Tübingen \quad
  $^6$Prior Labs \quad
}

\iclrfinalcopy

\begin{document}

\maketitle

\begin{abstract}
Symbolic regression is the problem of finding an expression that fits a dataset. Solving it requires searching a combinatorial space of candidate expressions.
Neural symbolic regression can amortize that search by training a model once on synthetic data and then predicting expressions for a new dataset. But existing models still underperform genetic programming on real, noisy data.
We argue that the deficit lies in the training rather than in the paradigm, and present \emph{RASR} (Reinforced Amortized Symbolic Regression), a method with two key changes to the training recipe. 
First, we design a synthetic prior that narrows the gap with real-world data: inputs come from the prior of a tabular foundation model; constants are sampled bottom-up, conditioned on the data; targets carry heteroscedastic and structured noise; and everything is standardized.
Second, we split training into two stages. Supervised pre-training maximizes the token-level likelihood of a reference expression. This is followed by reinforcement-learning post-training that aligns the model with its inference-time use by rewarding held-out fit and penalizing expression size.
On the black-box track of SRBench~2.0, RASR achieves the best mean test $R^2$ rank of all evaluated methods.
Three RASR variants, trained with different size penalties, make up most of the accuracy-complexity Pareto front. 
On Strogatz, our method recovers more ground-truth expressions than any baseline, and on Feynman it is competitive with the best-performing methods.

\end{abstract}

\section{Introduction}\label{sec:introduction}

\begin{figure}[tbp]
  \centering
  \includegraphics[width=0.81\linewidth]{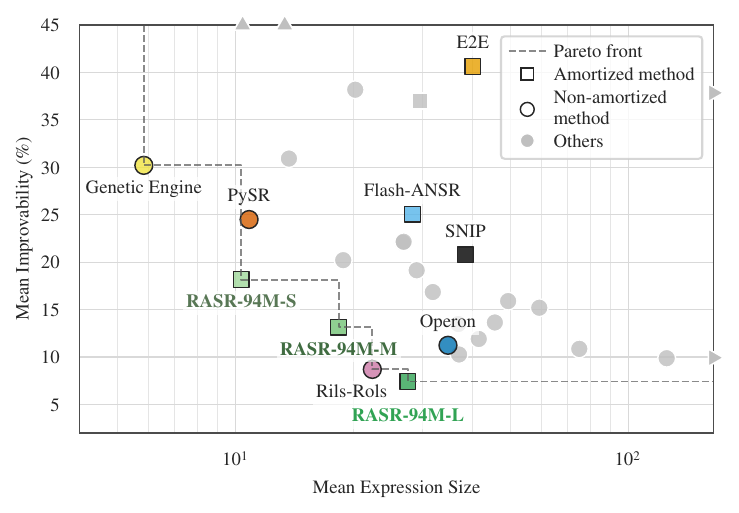}
  \caption{\textbf{Accuracy-complexity trade-off on the black-box track of SRBench~2.0.} Mean
  expression size (log scale) against mean improvability, the relative RMSE gap to the best method on each
  dataset (\%, lower is better). The dashed line marks the Pareto front, and triangles on the
  border mark methods outside the plotted range. Three of the five
  methods on the front are our RASR variants, differing only in the expression-size penalty used during
  post-training, and the default RASR-94M-L attains the best fit of any evaluated method.  \figref{fig:app-pareto-full} shows every method.
}
  \label{fig:blackbox-pareto}
\end{figure}

Symbolic regression is the task of finding a closed-form mathematical expression that fits a
regression dataset. The classical use case is scientific discovery, where the expression itself is
the object of interest~\citep{schmidt2009distilling}. Another is general tabular
prediction, where the data need not follow an exact law and symbolic regression can
still yield models that are interpretable yet expressive. Searching over expressions is difficult, since the space is discrete and
grows combinatorially with expression size, and the problem is NP-hard in
general~\citep{DBLP:journals/tmlr/VirgolinP22}. For decades this search has been driven
predominantly by genetic programming~\citep{koza1994genetic}, which is still widely regarded as the
state of the art, with PySR~\citep{DBLP:journals/corr/abs-2305-01582} and
Operon~\citep{DBLP:conf/gecco/BurlacuKK20} among its strongest modern representatives.

A more recent line of work is neural symbolic regression, where a transformer is trained over many synthetic
examples and then predicts expressions for a new dataset directly~\citep{DBLP:conf/icml/BiggioBNLP21,DBLP:conf/nips/KamiennydLC22}, amortizing a search that
classical methods must rerun for every problem. We use the term in this narrow sense of amortized
methods, excluding neural methods that train anew on each dataset (\secref{sec:relatedwork}). A similar recipe has reshaped the neighboring
field of tabular prediction, where tabular foundation models such as
TabPFN~\citep{DBLP:conf/iclr/Hollmann0EH23,DBLP:journals/nature/HollmannMPKKHSH25}, pre-trained
once on a synthetic prior, now match or surpass classical
approaches~\citep{DBLP:journals/corr/abs-2506-16791}. Neural symbolic regression, by contrast,
still trails the best search-based systems, particularly on black-box
problems~\citep{DBLP:conf/nips/CavaOBFVJKM21,DBLP:conf/gecco/AldeiaZBCFBCF25}. We argue that this gap
is not inherent to the paradigm but stems from how these models are trained.

Two aspects of that training are at fault. The first is the synthetic prior, the distribution of
training samples. Prior work samples inputs from uniform or Gaussian distributions and computes
targets without noise, as in NeSymReS and
E2E~\citep{DBLP:conf/icml/BiggioBNLP21,DBLP:conf/nips/KamiennydLC22}, so the tables a model learns
from look little like the ones it is applied to. The second is the training objective. Existing methods train almost exclusively with supervised objectives against a reference expression, most commonly token-level cross-entropy. At test time, however, we care more about fit quality and expression size than about the exact token sequence.

We address both in RASR, short for Reinforced Amortized Symbolic Regression.\footnote{
Our source code is publicly available at \url{https://github.com/automl/rasr}. The weights of our models are hosted on Hugging Face at \url{https://hf.co/collections/davidotte/rasr}.
}
At its core is an
encoder-decoder transformer that reads a dataset and generates expressions for it. At inference
RASR samples many candidates, refines their constants numerically, and returns the Pareto front of
fit quality against expression size, so a user can choose the trade-off. Our main contributions are:

\begin{itemize}
  \item \textbf{A diverse, data-realistic synthetic prior} whose inputs come from the prior of a
  tabular foundation model~\citep{DBLP:journals/corr/abs-2602-11139}, whose constants are sampled
  bottom-up conditioned on the data so that every operator stays in a sensible regime, and whose
  targets carry heteroscedastic and structured noise, with full standardization throughout
  (\secref{sec:prior}).
  \item \textbf{A two-stage training procedure} that keeps supervised pre-training as the scalable
  stage and adds a reinforcement-learning post-training stage, rewarding held-out fit and
  penalizing expression size directly with group-relative policy
  optimization~\citep{DBLP:journals/corr/abs-2402-03300} under a risk-sensitive
  variant~\citep{DBLP:journals/corr/abs-2509-24261} (\secref{sec:training}).
  \item \textbf{State-of-the-art results on SRBench~2.0}, with the best mean test $R^2$ rank of
  all evaluated methods on the black-box track, at modest expression size. Three variants,
  post-trained with different size penalties, span the accuracy-complexity trade-off and all lie
  on the Pareto front, which holds only two other methods (\figref{fig:blackbox-pareto}). On the ground-truth
  tracks of the original SRBench, RASR attains the highest solution rate of all methods on
  Strogatz and is close behind the leaders on Feynman (\secref{sec:results}).
\end{itemize}

\section{Related Work}\label{sec:relatedwork}

Symbolic regression is classically approached with genetic programming, which evolves a population
of candidate expression trees through fitness-based selection and tree-level mutation and crossover
operators~\citep{koza1994genetic}, with PySR~\citep{DBLP:journals/corr/abs-2305-01582} and
Operon~\citep{DBLP:conf/gecco/BurlacuKK20} as widely used implementations. Our method belongs to neural symbolic regression, which trains a
sequence model once on large amounts of synthetic dataset-expression
pairs~\citep{DBLP:conf/icml/BiggioBNLP21,DBLP:conf/nips/KamiennydLC22,DBLP:journals/corr/abs-2602-08885}.
Other neural approaches train anew on each dataset, such as EQL~\citep{DBLP:conf/icml/SahooLM18},
which fits a network with symbolic activation functions, and deep symbolic regression
(DSR)~\citep{DBLP:conf/iclr/PetersenLMSKK21}, which learns a generative policy over expressions
from a fit-based reward.

\textbf{Four design choices.} Nearly all neural symbolic regression methods follow one recipe, that is, train
once on synthetic dataset-expression pairs, then encode a given dataset and decode expressions for
it. They differ on four recurring choices. The first is the synthetic prior. Expressions are
sampled with the tree-counting strategy of \citet{DBLP:conf/iclr/LampleC20}, while the accompanying
datasets are built from simple input distributions without target noise. The second is the training objective, usually a
single stage of token-level cross-entropy, for which richer multimodal alternatives have been
proposed~\citep{DBLP:conf/iclr/MeidaniSRF24,DBLP:conf/iclr/LiLSWYLLT23}, including Sym-Q~\citep{tian2025interactive}. The
third is the treatment of numerical constants, which are either left as placeholders for a
downstream optimizer or predicted in high precision. The fourth is how test-time compute buys
candidates, through sampling~\citep{DBLP:conf/nips/KamiennydLC22}, beam
search~\citep{DBLP:conf/icml/BiggioBNLP21}, a latent-space search~\citep{DBLP:conf/iclr/MeidaniSRF24}
or TPSR's Monte-Carlo tree search around a frozen decoder~\citep{DBLP:conf/nips/ShojaeeMFR23}.

\textbf{Positioning of RASR.} RASR departs from the usual answer on each of the four choices.
The first two carry our main contributions, the prior of \secref{sec:prior} and the
post-training stage of \secref{sec:training}. Reinforcement learning has been applied to symbolic regression before, but per dataset (DSR),
as a test-time search around a frozen decoder (TPSR), or as an offline objective folded into
pre-training itself (Sym-Q), so post-training a general pre-trained model against a task reward is
new to the field. On the remaining two choices our departures are smaller. We emit rounded
constants that act as an initialization region for a downstream optimizer, and we sample candidates
independently at test time instead of running a guided search, since post-training has already shifted
the policy toward strong candidates. \appref{app:relatedwork} treats each choice in full.
\section{Method}\label{sec:method}

In symbolic regression we are given a dataset
$\mathcal{D} = \{(\bm{x}_i, y_i)\}_{i=1}^{n}$ with inputs $\bm{x}_i \in \mathbb{R}^{d}$ and
targets $y_i \in \mathbb{R}$, and seek a closed-form expression $f$ that both fits the data and
stays compact. At the core of RASR is a transformer mapping $\mathcal{D}$ to a distribution over such
expressions. We describe its four ingredients in turn: the synthetic prior the model is trained on
(\secref{sec:prior}), the architecture (\secref{sec:architecture}), the two-stage training
procedure (\secref{sec:training}), and the inference pipeline that refines and selects among the
proposed expressions (\secref{sec:inference}).

\subsection{Synthetic Prior}\label{sec:prior}

\begin{figure}[tbp]
  \centering
  \resizebox{0.82\linewidth}{!}{\input{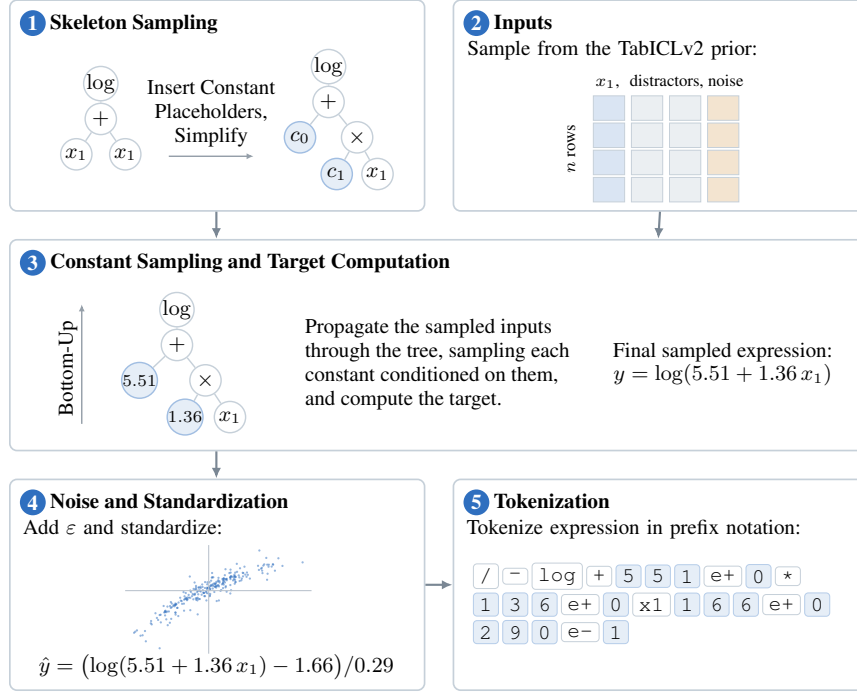}}
  \caption{\textbf{Overview of the synthetic training data generation.} Each training pair is produced by sampling an
  expression skeleton, drawing the inputs together with an auxiliary noise node from the TabICLv2
  regression prior, filling the constant placeholders bottom-up conditioned on those inputs, and
  optionally injecting noise before standardizing and tokenizing. One skeleton is followed
  throughout: its constants, the target panel of stage~4 and the token row all come from a single
  draw through the pipeline.}
  \label{fig:prior}
\end{figure}

Each pre-training example is a pair $(\mathcal{D}, e)$, where $e$ is a ground-truth expression used
during the dataset's generation, and the task is to recover $e$ from $\mathcal{D}$. The quality
of the prior these pairs are drawn from largely determines how well the model transfers to real
data, and is therefore a central design component of this work. Sampling proceeds in the stages
of \figref{fig:prior}, and full details and hyperparameters are given in \appref{app:prior}.

\textbf{Skeleton sampling.} Expressions are built from three binary and fifteen unary
operators over variables $x_1,\dots,x_{11}$, all listed in \appref{app:skeleton}. We sample the number of distinct variables $n_v$ uniformly from
$\{1,\dots,11\}$, then draw a binary tree from the unbiased tree-counting distribution of
\citet{DBLP:conf/iclr/LampleC20} and fill its leaves with variables. Following
E2E~\citep{DBLP:conf/nips/KamiennydLC22}, the unary operators are inserted afterwards at uniformly
drawn positions, which keeps the number of binary and unary operators under independent control.
We then insert constant placeholders wherever a constant would meaningfully change the function,
omitting those a later standardization or a parent constant would make redundant, and simplify
the result with SimpliPy~\citep{DBLP:journals/corr/abs-2602-08885}. The principle is
to make every operator argument freely scalable and shiftable, so the sampler can place each
sub-function in a sensible regime: the argument of every unary operator
$g$ is wrapped affinely, $g(t) \rightarrow g(c_0 t + c_1)$, binary operators receive an additive
bias, and variable leaves and the atomic terms of a sum a multiplicative scale. Inserting many placeholders is
deliberate: since the model only sees standardized inputs on a bounded domain, these constants
are the primary lever by which one skeleton realizes qualitatively different functions.

\textbf{Input sampling.} Previous work samples the input matrix $X$ from simple fixed
distributions. We instead draw it from a tabular foundation model prior, so that the features
and their relationships resemble those of real tabular data. Such priors sample a random
structural causal model, a directed acyclic graph over variables with a random function at each
node, and read a subset of the resulting variables off as
columns~\citep{DBLP:conf/iclr/Hollmann0EH23}. Because all columns arise from the same coupled
mechanism, they can be nonlinearly related, correlated and redundant, as they are in real data.
We use the prior of TabICLv2~\citep{DBLP:journals/corr/abs-2602-11139}, drawing a
pool of twelve standardized numerical columns: the first $n_v$ become the expression's variables, one is held
out as a noise node shaping input-correlated noise, and a random number of the rest are appended
to $\mathcal{D}$ as correlated distractor features absent from $f$, so the model must learn to
ignore irrelevant inputs. The number of observations $n$ is sampled uniformly from $8$ to
$1024$. We use the prior essentially as published, with targeted modifications that make its
samples better suited as continuous inputs (\appref{app:tabicl}).

\textbf{Bottom-up constant sampling.} Filling the placeholders independently from a fixed
distribution has two problems. It produces degenerate functions, for example arguments that
saturate $\exp$ or fall outside the domain of $\log$, and it does not ensure the expression is
one a user would consider a reasonable solution given the dataset, as characteristic shapes may
be missed entirely or one term may dominate another by orders of magnitude. We therefore sample
constants conditioned on the data in a single bottom-up pass over the tree: each subtree is
evaluated on $X$, and when the evaluation reaches an affine wrapper $c_0 t + c_1$, the constants
are drawn conditioned on the empirical distribution of the child values $t$, so that the operator receives
inputs in a sensible regime by construction. Positive-domain operators keep their argument's
minimum above zero, and the terms of a sum are scaled jointly so
that none vanishes or dominates (\appref{app:constants}).

\textbf{Noise and standardization.} Real data is rarely clean, and a model trained only on
clean targets may fail to generalize. We inject noise into most datasets while leaving $40\%$
untouched, at a strength drawn per dataset from an exponential distribution in units of the clean
target's standard deviation, so most datasets are only mildly disturbed with a long tail of
stronger corruption. Two forms are used equally often. The first is Gaussian, with a per-row
standard deviation that blends a constant with a profile derived from the auxiliary noise node.
Since that node's relation to the observed features is inherited from the SCM prior, the noise
spans homoscedastic through input-correlated heteroscedastic while staying zero-mean at each row,
so the ground truth remains the conditional mean $\mathbb{E}[y \mid \bm{x}]$ and hence the optimal
predictor. The second form instead mimics unobserved confounders: the noise node is passed
through a random activation and centered over the dataset, which adds structured deviations
rather than white noise. Since it is zero-mean only marginally, and not conditionally on
$\bm{x}$, the ground-truth expression is here an expressible fit rather than the exact
conditional mean (\appref{app:dataset}). Finally, since real targets arrive at arbitrary scales,
we standardize the target as $\hat{y} = (y - \mu_y)/\sigma_y$ and rewrite the expression to
describe $\hat{y}$ rather than $y$.

\textbf{Tokenization and constant strategy.} Expressions are serialized in prefix notation,
where operators and variables map to single tokens, while numeric constants are emitted as digit
tokens, a mantissa rounded to three significant figures with a single-digit exponent. We treat
these values only as an initialization for the refinement of \secref{sec:inference}, so the model
learns an approximate region of constant space rather than an exact value.

\subsection{Architecture}\label{sec:architecture}

The model is an encoder-decoder transformer that maps an input table to symbolic expressions. The
encoder turns the table, whose first column holds the target, into one latent per row, and the
decoder generates an expression as a prefix token sequence while cross-attending to those latents,
so sampling it repeatedly yields a set of candidates. We deliberately keep the backbone close to a standard encoder-decoder
transformer~\citep{DBLP:conf/nips/VaswaniSPUJGKP17} and apply only targeted changes.

Every scalar cell is embedded through the raw IEEE-754 bits of its \texttt{float32}
value~\citep{DBLP:conf/icml/BiggioBNLP21}, giving the model direct access to magnitude and sign,
and the encoder alternates attention over the two axes of the table, following
TabPFNv2~\citep{DBLP:journals/nature/HollmannMPKKHSH25}. Since sample-axis attention treats the
rows as an unordered set, the encoder is permutation-equivariant over samples and accepts a
variable number of rows, and each row's target-column representation becomes its latent. The
number of rows varies by two orders of magnitude across datasets, so along that axis we replace
the usual fixed attention scale with the length-robust
QASSMax~\citep{DBLP:journals/corr/abs-2602-11139}. Further changes are gated
attention~\citep{DBLP:conf/nips/QiuWZHWYMYHHLZL25}, and QK-norm~\citep{DBLP:conf/emnlp/HenryDPC20}
and pre-normalization with RMSNorm~\citep{DBLP:conf/nips/ZhangS19a} for training stability. Our
main model has $94$M parameters, alongside a $24$M variant for ablations.
\appref{app:architecture} gives the full description and a diagram.

\subsection{Training}\label{sec:training}

\begin{figure}[tbp]
  \centering
  \resizebox{0.95\linewidth}{!}{
%
{%
\definecolor{prioraccent}{HTML}{2F6FBF}%
\definecolor{priorwarm}{HTML}{C9852A}%
\definecolor{priorsurface}{HTML}{F7F9FB}%
\definecolor{priorborder}{HTML}{C6D0D9}%
\definecolor{priormuted}{HTML}{7A8794}%
\rmfamily
\begin{tikzpicture}[
    every node/.style={text=black, font=\small},
    hrule/.style={draw=priorborder, line width=0.5pt},
    divider/.style={draw=priorborder, line width=0.7pt},
    badge/.style={
      circle, fill=prioraccent, text=white, font=\bfseries\small,
      inner sep=0pt, minimum size=3.8mm,
    },
    title/.style={font=\fontsize{9.6}{11.5}\selectfont\bfseries, inner sep=0pt},
    subtitle/.style={font=\fontsize{9.6}{11.5}\selectfont, inner sep=0pt},
    chip/.style={
      rounded corners=1.5pt, draw=priorborder, fill=white, font=\small\ttfamily,
      inner xsep=2.8pt, inner ysep=1pt, text=black,
      text height=1.25ex, text depth=0.2ex,
    },
    numchip/.style={chip, fill=white, draw=prioraccent!35},
    wrongchip/.style={draw=red!60},
    modelbox/.style={
      draw=prioraccent!45, line width=0.7pt, rounded corners=2.5pt,
      fill=prioraccent!12, font=\bfseries\small, inner sep=4pt,
      align=center, text=black,
    },
    flow/.style={-{Latex[length=4.5pt,width=4pt]}, draw=priormuted,
                 line width=0.9pt, rounded corners=2pt},
    flowlabel/.style={font=\small},
    loss/.style={
      draw=priorwarm!70!black, line width=0.6pt, rounded corners=2.5pt,
      fill=priorwarm!18, align=center, inner sep=5pt, font=\bfseries\small,
    },
    cell/.style={draw=priorborder, line width=0.4pt, fill=white,
                 minimum width=3.6mm, minimum height=3.2mm, inner sep=0pt},
    targetcell/.style={cell, fill=prioraccent!25, draw=prioraccent!55},
    holdoutcell/.style={cell, fill=priorwarm!25, draw=priorwarm!55},
    bigdot/.style={circle, fill=priormuted, inner sep=0pt, minimum size=2.2pt},
    zone/.style={draw=priormuted, dashed, line width=0.6pt, rounded corners=3pt},
    zonelabel/.style={font=\bfseries\small, fill=white,
                       inner sep=1pt},
  ]

  \node[badge] (prebadge) at (0.5,6.6) {1};
  \node[title, anchor=west] at (0.75,6.6) {Pre-training};
  \node[subtitle, anchor=west] at (0.3,6.25) {Supervised, teacher forcing};

  \begin{scope}[xshift=0.71cm]
  \node[badge] (postbadge) at (8.05,6.6) {2};
  \node[title, anchor=west] at (8.30,6.6) {Post-training};
  \node[subtitle, anchor=west] at (7.80,6.25)
    {Reinforcement learning, no ground-truth expression};
  \end{scope}

  \draw[divider] (8.005,-0.10) -- (8.005,7.0);

  \foreach \c/\style in {0/targetcell, 1/cell, 2/cell} {
    \foreach \r in {0,1,2} {
      \node[\style] at (0.7+0.4*\c, 1.0+0.36*\r) {};
    }
  }
  \node[font=\small, anchor=north] at (1.1,0.80) {Dataset $\mathcal{D}$};
  \coordinate (pretbltop) at (1.1,1.90);

  \node[modelbox, minimum width=1.7cm, minimum height=0.85cm] (preenc) at (1.1,3.8) {Encoder};
  \draw[flow] (pretbltop) -- (preenc.south);

  \node[modelbox, minimum width=2.6cm, minimum height=0.85cm] (predec) at (3.65,3.8) {Decoder};
  \draw[flow] (preenc.east) -- (predec.west);

  \node[chip, inner xsep=1.8pt] (gt1) at (2.50,2.4) {/};
  \node[chip, inner xsep=1.8pt] (gt2) at (2.90,2.4) {$-$};
  \node[chip, inner xsep=1.8pt] (gt3) at (3.45,2.4) {sin};
  \node[chip, inner xsep=1.8pt] (gt4) at (4.00,2.4) {$-$};
  \node[numchip, inner xsep=1.8pt] (gt5) at (4.40,2.4) {3};
  \node[numchip, inner xsep=1.8pt] (gt6) at (4.80,2.4) {9};
  \foreach \dx in {0.16,0.29,0.42} { \node[bigdot] at ($(gt6.east)+(\dx,0)$) {}; }
  \node[font=\small, anchor=north] at (3.65,2.15) {Ground truth $e$};
  \draw[flow] (gt1.north) -- (2.50,3.375);
  \draw[flow] (gt2.north) -- (2.90,3.375);
  \draw[flow] (gt3.north) -- (3.45,3.375);
  \draw[flow] (gt4.north) -- (4.00,3.375);
  \draw[flow] (gt5.north) -- (4.40,3.375);
  \draw[flow] (gt6.north) -- (4.80,3.375);

  \node[chip, inner xsep=1.8pt] (pr1) at (2.50,5.2) {$-$};
  \node[chip, inner xsep=1.8pt] (pr2) at (3.05,5.2) {sin};
  \node[chip, wrongchip, inner xsep=1.8pt] (pr3) at (3.60,5.2) {$+$};
  \node[numchip, inner xsep=1.8pt] (pr4) at (4.00,5.2) {3};
  \node[numchip, wrongchip, inner xsep=1.8pt] (pr5) at (4.40,5.2) {7};
  \node[numchip, inner xsep=1.8pt] (pr6) at (4.80,5.2) {1};
  \foreach \dx in {0.16,0.29,0.42} { \node[bigdot] at ($(pr6.east)+(\dx,0)$) {}; }
  \node[font=\small, anchor=south] at (3.65,5.45) {Next-token predictions};
  \draw[flow] (2.50,4.225) -- (pr1.south);
  \draw[flow] (3.05,4.225) -- (pr2.south);
  \draw[flow] (3.60,4.225) -- (pr3.south);
  \draw[flow] (4.00,4.225) -- (pr4.south);
  \draw[flow] (4.40,4.225) -- (pr5.south);
  \draw[flow] (4.80,4.225) -- (pr6.south);

  \node[loss, minimum width=2.0cm, minimum height=0.85cm] (ce) at (6.35,3.8) {Cross-Entropy\\Loss};
  \draw[flow] ($(pr6.east)+(0.52,0)$) -| (ce.north);
  \draw[flow] ($(gt6.east)+(0.52,0)$) -| (ce.south);

  \begin{scope}[xshift=0.71cm]
  \foreach \c/\style in {0/targetcell, 1/cell, 2/cell} {
    \foreach \r in {0,1,2} { \node[\style] at (8.65+0.4*\c, 1.60+0.36*\r) {}; }
  }
  \foreach \c/\style in {0/holdoutcell, 1/cell, 2/cell} {
    \foreach \r in {0,1} { \node[\style] at (8.65+0.4*\c, 0.35+0.36*\r) {}; }
  }
  \node[font=\small, anchor=east, align=right] (trainlab)
    at (8.45,2.06) {Train\\split};
  \node[font=\small, anchor=east, align=right] (testlab)
    at (8.45,0.53) {Test\\split};
  \coordinate (posttbltop) at (8.65,2.50);
  \coordinate (traintblright) at (9.63,2.06);
  \coordinate (testtblright) at (9.63,0.53);

  \node[modelbox, minimum width=1.7cm, minimum height=0.85cm] (postenc) at (8.65,4.25) {Encoder};
  \draw[flow] (posttbltop) -- (postenc.south);

  \node[modelbox, minimum width=2.1cm, minimum height=0.95cm] (postdec)
    at (11.15,4.25) {Decoder};
  \draw[flow] (postenc.east) -- (postdec.west);
  \draw[flow] ($(postdec.north)+(-0.35cm,0)$) .. controls
    ++(-0.25cm,0.9cm) and ++(0.25cm,0.9cm) .. ($(postdec.north)+(0.35cm,0)$);
  \node[flowlabel, align=center] at (11.60,5.62)
    {Autoregressive rollout};

  \node[anchor=west, font=\bfseries\small] (c1lab) at (12.55,4.55) {1.};
  \node[chip, anchor=west, right=0.3pt of c1lab] (c1) {$2.13\,x_1 - 1.08$};
  \node[anchor=west, font=\bfseries\small] (c2lab) at (12.55,4.15) {2.};
  \node[chip, anchor=west, right=0.3pt of c2lab] (c2) {$0.51\,\sin(0.3 x_2)$};
  \draw[flow] (postdec.east) -- (12.55,4.25);

  \node[modelbox, fill=white, minimum width=2.1cm, minimum height=0.95cm] (refine)
    at (13.70,2.06) {Refine Constants};

  \foreach \delta in {3.80,3.68,3.56} { \node[bigdot] at (13.70,\delta) {}; }
  \draw[flow] (13.70,3.41) -- (refine.north);
  \draw[flow] (traintblright) -- (refine.west);

  \node[loss, minimum width=2.35cm, align=center] (rew) at (13.70,0.53)
    {\textbf{Reward:} \mdseries\small Based on\\[1pt] \mdseries\small fit accuracy \& size};
  \draw[flow] (refine.south) -- (rew.north);
  \draw[flow] (testtblright) -- (rew.west);

  \draw[zone] (10.10,1.40) rectangle (15.35,5.80);
  \node[zonelabel, anchor=south] at (12.73,5.85) {Inference};
  \end{scope}

\end{tikzpicture}%
}}
  \vspace{2pt}
  \caption{\textbf{The two-stage training process of RASR.} Pre-training maximizes the likelihood of the
  ground-truth prefix sequence of each synthetic pair under teacher forcing. Post-training
  discards the ground-truth expression entirely: the policy samples a group of candidates per
  dataset, each candidate's constants are refined on a training split and scored on a held-out
  split, and the resulting fit and size form a reward turned into a risk-sensitive group-relative
  advantage.}
  \label{fig:training}
\end{figure}
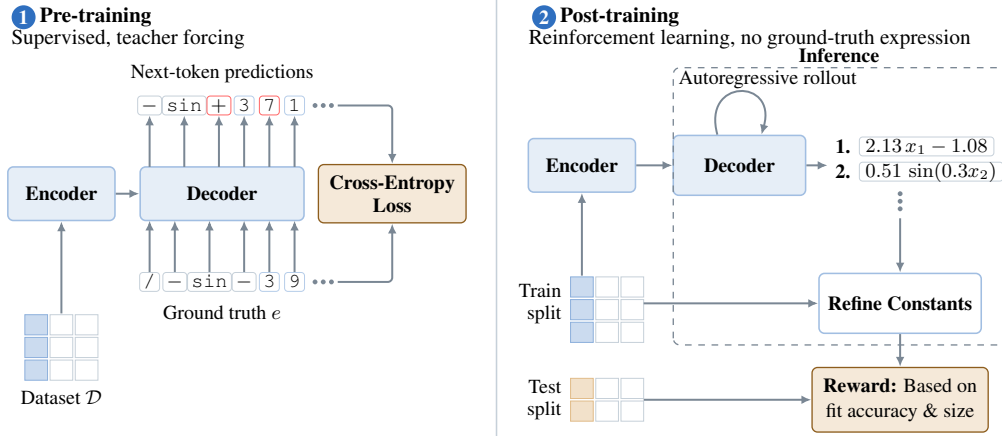

The model is trained in the two stages of \figref{fig:training}, which serve distinct purposes.
Pre-training is where scale happens: a cheap, massively parallel supervised objective over the
synthetic pairs of \secref{sec:prior}, whose role is to train on as much data as the budget allows
and leave a strong initialization. Post-training then aligns that initialization with how the
model is actually used, replacing the surrogate objective with the quantities we measure at
inference, and teacher forcing with the model's own autoregressive samples.

\textbf{Pre-training.} Given the encoded dataset, the decoder is trained to predict the
ground-truth prefix expression one token at a time under a token-level cross-entropy objective,
with each step conditioned on the true prefix rather than on the model's own output (teacher
forcing). Pairs are generated offline and streamed, so each is seen at most once. The $94$M model
sees roughly $62$M datasets over $242$k steps (\appref{app:training}).

\textbf{Post-training.} A subsequent reinforcement-learning stage addresses several mismatches
between pre-training and test time. Pre-training targets token-level agreement with one reference
expression while at test time fit is what matters the most, and since many expressions describe
the same function, token-level matching is a loose surrogate at best. Teacher forcing conditions
the model on ground-truth prefixes rather than its own, a mismatch known as exposure bias, where
early errors compound once it decodes alone. Finally, inference draws many candidates and returns
the best, so post-training can absorb part of that search into the model itself.

\textbf{Policy optimization.} Fit can only be measured on sampled autoregressive rollouts, which
are not differentiable, so we treat the model as a policy over expressions and update it with group-relative policy optimization
(GRPO)~\citep{DBLP:journals/corr/abs-2402-03300}. GRPO estimates its baseline from a group of $G$
samples drawn per input rather than from a learned value network, so a sample's advantage is its
reward $r_i$ standardized within the group,
$\hat{A}_i = (r_i - \operatorname{mean}(r))/\operatorname{std}(r)$. Because we care about the best
of many samples rather than the average one, we take the advantage of
RS-GRPO~\citep{DBLP:journals/corr/abs-2509-24261} unchanged in its place,
$\hat{A}^{\text{RS}}_i = \beta^{-1}\big(e^{\beta r_i} / \tfrac{1}{G}\sum_{j} e^{\beta r_j} - 1\big)$,
which arises from replacing the risk-neutral expected reward with an exponential utility and leaves
the rest of the GRPO machinery intact. It weights the update exponentially towards the
highest-reward samples in each group and, across a batch, shifts learning towards the harder
inputs. A positive $\beta$ thus tilts the objective towards the best-of-$n$ target that
sampling-based inference actually measures, while $\beta \to 0$ recovers ordinary GRPO. We use
$\beta = 4$. \appref{app:grpo} states both objectives in full.

\textbf{Reward.} We design the reward $R(e)$ of a completed expression $e$ along two axes: fit,
for which we adopt the normalized RMSE reward of DSR~\citep{DBLP:conf/iclr/PetersenLMSKK21}, and
expression size, a penalty on the number of nodes $\mathcal{C}(e)$ in the expression tree. Treating the
two as separate objectives and standardizing each within the group, written
$\hat{R}_{\text{fit}}$ and $\hat{\mathcal{C}}$, gives
\begin{equation}
  R(e) \;=\; \hat{R}_{\text{fit}}(e) \;-\; \lambda\, \hat{\mathcal{C}}(e),
  \qquad
  R_{\text{fit}}(e) \;=\; \frac{1}{1 + \text{NRMSE}(e)} \;\in\; (0, 1],
  \label{eq:reward}
\end{equation}
with a trade-off weight $\lambda \geq 0$. The normalized error
$\text{NRMSE} = \sigma_y^{-1}\sqrt{n^{-1}\sum_i (y_i - \hat{y}_i)^2}$ is scale-free and therefore
comparable across datasets. Crucially, it is evaluated on a held-out split: the constants are
refined on the training split and the fit computed on a split the optimizer never saw, which
rewards genuine generalization rather than memorization of the fitted points. Invalid expressions
and failed fits receive a fit reward of zero.

\textbf{Data and variants.} Post-training needs no ground-truth expression, since the reward
depends only on the dataset and the predictions. We target both the black-box setting, where the
data is not guaranteed to come from a closed-form expression, and the classical setting, where it
is. Post-training therefore uses an equal mixture of freshly sampled datasets from our pre-training
prior and datasets drawn directly from the modified TabICLv2 prior, whose underlying function need
not be expressible in our primitive set. We sample $128$ generations per dataset at temperature
$1.0$ and disable the KL penalty entirely. The weight $\lambda$ has no single correct setting,
since how much fit a user will give up for a smaller expression depends on the application, so we
treat it as a property of the trained model and post-train three variants of the $94$M model from
the same checkpoint, at $\lambda \in \{0.05, 0.15, 0.5\}$ with $\lambda = 0.05$ as the default.
Applying the penalty during post-training rather than at selection time shapes the sampling
distribution itself, so a strongly penalized variant searches within a compact region of expression
space instead of filtering a pool of large candidates later.

\subsection{Inference}\label{sec:inference}

At test time we encode $\mathcal{D}$ once and autoregressively sample $K$ candidates, discarding
those that are invalid, too long, or reference more variables than the
dataset provides. Each remaining unique candidate's constants are refined by nonlinear least
squares, using the Levenberg--Marquardt~\citep{levenberg1944method,marquardt1963algorithm} solver
in \texttt{scipy}~\citep{DBLP:journals/corr/abs-1907-10121} seeded from the emitted values with
several Gaussian-perturbed restarts. Candidates are scored on fit, measured by $R^2$, and size
$\mathcal{C}(e)$, and exposed in two ways: whenever a user can make the trade-off themselves we
return the Pareto front, the candidates not beaten on both axes at once, and when a single
expression must be returned without a human in the loop, as in a benchmark evaluation, we select
the candidate maximizing
\begin{equation}
  S(e) \;=\; R^2(e) \;-\; \gamma\, \mathcal{C}(e),
  \label{eq:selection}
\end{equation}
where the parsimony weight $\gamma \geq 0$ states how much additional fit a candidate must
deliver to justify one further node. \appref{app:inference} gives more details.

\section{Experiments}\label{sec:results}

We evaluate RASR on SRBench~2.0~\citep{DBLP:conf/gecco/AldeiaZBCFBCF25}, the recent successor to
the widely used SRBench suite~\citep{DBLP:conf/nips/CavaOBFVJKM21}, and on the Feynman and
Strogatz tracks of the original SRBench. This section gives an overview of the main results.
Additional results and the ablations are in the appendix, summarized in
\secref{subsec:further}.

\subsection{Setup}\label{subsec:setup}

\textbf{Benchmarks and protocol.} The black-box track of SRBench~2.0 is our primary point of
evaluation. Its twelve regression datasets come from the PMLB
collection~\citep{DBLP:journals/biodatamining/OlsonCOUM17}, mostly real-world
problems without a known closed-form target, curated to be harder and more representative than in the original SRBench suite. The other
tracks instead pair every problem with a known expression. We follow the respective evaluation
protocols. RASR samples $K = 1024$ candidates per run, from a single configuration left unchanged
across all datasets and tracks. The only quantity we tune is the scalar parsimony weight $\gamma
\in \{0, 0.01, 0.1\}$ of \eqref{eq:selection}, selected by cross-validation on the training split,
which acts at selection time on an already generated candidate pool rather than steering the
search. Since RASR was trained on at most $1024$ rows and $11$ features, we cap its input at $2048$
rows, twice the training maximum, and at $11$ features, which reduces four of the twelve black-box datasets and subsamples the Feynman problems.
\appref{app:datasets} describes the datasets of all four tracks and \appref{app:protocol} the
remaining protocol details.

\textbf{Metrics, baselines and variants.} We record each run's test $R^2$ and expression size,
counted as the number of nodes after conversion to SymPy~\citep{10.7717/peerj-cs.103}, and
aggregate a method's runs on a dataset by the median. On the
black-box track we summarize the per-dataset medians by two complementary track averages, the mean
test $R^2$ rank, which only depends on the ordering of methods, and the mean improvability~\citep{DBLP:journals/corr/abs-2506-16791}, $(\mathrm{err}_i - \mathrm{err}^\star_i)/\mathrm{err}_i$
with $\mathrm{err}_i$ a method's test RMSE on dataset $i$ and $\mathrm{err}^\star_i$ the lowest of a fixed
pool, which retains the magnitude of the gaps (\appref{app:protocol}). As baselines we take
the full set of methods evaluated by the benchmark that defines each track, covering
genetic-programming systems, non-evolutionary search and neural methods. To all four tracks we add
two pre-trained models that neither benchmark includes,
Flash-ANSR~\citep{DBLP:journals/corr/abs-2602-08885} and SNIP~\citep{DBLP:conf/iclr/MeidaniSRF24},
run under the same protocol. Our variants are named by parameter count and by the typical size of
the expressions they return, which the post-training penalty sets: RASR-94M-L ($\lambda = 0.05$) is the main model, RASR-94M-M ($\lambda = 0.15$) and RASR-94M-S ($\lambda = 0.5$) the
smaller-expression variants.

\subsection{Black-Box Track}\label{subsec:blackbox}

\begin{figure}[tbp]
  \centering
  \includegraphics[width=\linewidth]{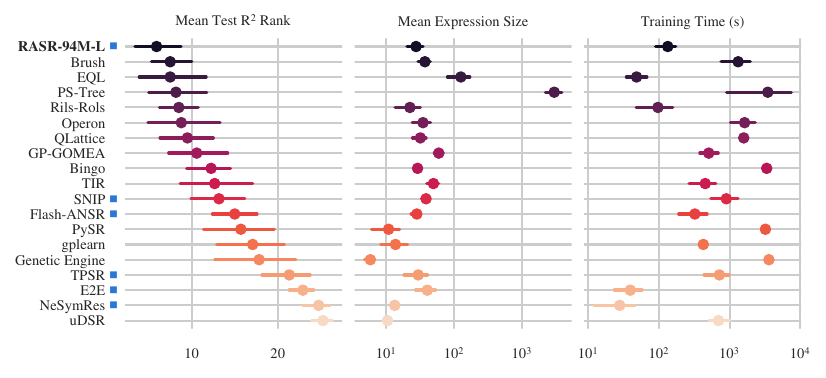}
  \caption{\textbf{Black-box track overview.} Methods are ordered by mean test $R^2$ rank across the twelve
  datasets (left, lower is better), while the middle and right panels show mean expression size and
  fitting time on log scales. Markers are means across datasets, bars bootstrapped $95\%$ confidence intervals. Amortized methods are flagged with a blue square next to their name. For
  readability only selected baselines are shown, with ranks computed against all baselines.
  \figref{fig:app-overview-variants} shows every method.}
  \label{fig:blackbox-overview}
\end{figure}

\figref{fig:blackbox-overview} orders the evaluated methods by mean test $R^2$ rank, with mean
expression size and fitting time alongside, so that accuracy is read together with the size and
cost at which it is bought. RASR-94M-L attains the best mean rank, ahead of Brush, EQL, PS-Tree and
Rils-Rols, at a mean expression size of $27$ nodes and with less fitting time than the
genetic-programming systems next to it. The methods ranking close behind pay for their
accuracy in size, EQL with roughly five times as many nodes and PS-Tree with two orders of
magnitude more. The other amortized methods sit far lower, the best of them, SNIP and Flash-ANSR, at ranks 18 and 19 of all 28 methods
(\figref{fig:app-overview-variants}). RASR is the first amortized method to close that gap.

\figref{fig:blackbox-pareto} shows the results as a trade-off, in the plane of mean expression
size against mean improvability. All three RASR variants lie on the Pareto front, RASR-94M-L anchoring the best-fit end at
$7.4\%$ improvability and the other two giving up fit for expressions of $18$ and $10$ nodes. Only
Rils-Rols and Genetic Engine reach the front at all, so three of its five methods are ours. The
size penalty applied during post-training thus moves the model along the trade-off without pushing
it off the front, letting one recipe serve both ends of it. The same plot with every baseline
identified individually, rather than only the handful named here, is shown in
\appref{app:additional-blackbox}. When we evaluate only on the eight datasets within our bounds,
RASR-94M-L improves to roughly $4\%$ improvability against just under $10\%$ for the next baseline
(\appref{app:additional-blackbox}).

\subsection{Feynman and Strogatz}\label{subsec:feynman}

\begin{figure}[tbp]
  \centering
  \begin{minipage}[t]{0.498\linewidth}
    \centering
    \includegraphics[width=\linewidth]{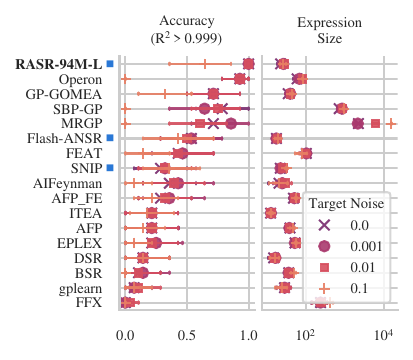}
  \end{minipage}\hfill
  \begin{minipage}[t]{0.498\linewidth}
    \centering
    \includegraphics[width=\linewidth]{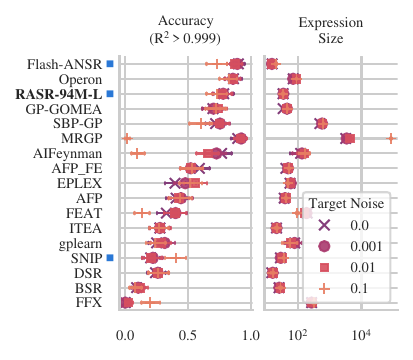}
  \end{minipage}
  \vspace{-2.5em}
  \caption{\textbf{Ground-truth recovery on the Strogatz (left) and Feynman (right) tracks.} Panels give
  the accuracy solution rate, the fraction of runs reaching test $R^2 > 0.999$
  (higher is better), and the mean expression size on a log scale (lower is better). Markers
  show target-noise levels with $95\%$ confidence intervals over problems. Methods are ordered by
  solution rate averaged over noise levels. Amortized methods are flagged with a blue square next to
  their name.}
  \label{fig:groundtruth}
\end{figure}

Feynman and Strogatz generate their problems from known laws. \figref{fig:groundtruth} scores both by the
accuracy solution rate, the fraction of runs reaching test $R^2 > 0.999$, at four different target-noise
levels. Without noise, RASR-94M-L solves
$99.3\%$ of the Strogatz problems, ahead of every other method and of Operon's $86.4\%$ at a third
of its expression size. On Feynman it reaches $78.9\%$ against $89.5\%$ for Flash-ANSR and
$85.2\%$ for Operon, a gap about half of which comes from the fourteen problems with six or more
variables, where it often falls slightly below the $0.999$ threshold (\appref{app:groundtruth}).
At $10\%$ target noise RASR still solves $65.0\%$ on Strogatz while Flash-ANSR drops to $41.4\%$
and Operon to $5.0\%$, and on Feynman it overtakes Flash-ANSR, at $72.5\%$ against $70.5\%$. Flash-ANSR was pre-trained on
a prior close to Feynman's independently and uniformly drawn inputs, but on Strogatz it is less
robust. \appref{app:groundtruth} gives the test $R^2$ view of the same runs.

\subsection{Additional Results and Ablations}\label{subsec:further}

On the first-principles track of SRBench~2.0, whose twelve small problems pair real or noisy data
with an accepted law, at least one RASR variant lies on the accuracy-size Pareto front for ten of
the twelve problems (\appref{app:firstprinciples}). \appref{app:posttraining-analysis} analyzes
post-training and \appref{app:ablations} ablates our changes on the black-box track, both in mean
improvability against a fixed pool of all baselines and RASR variants. At the $24$M scale the
ablations use, a model trained on a prior in the style of earlier neural symbolic-regression work
reaches $18.31$ after post-training, against $9.51$ for ours. Post-training improves all three
axes at the $94$M scale: it lowers improvability from $10.03$ to $7.44$, shortens the
returned expressions from $37.4$ to $27.5$ nodes and cuts the fitting time from $762$ to $133$
seconds. Improvability falls steadily as more candidate expressions are sampled, and
post-training reduces the number of samples needed.
Predicting constants as an initialization matters as well, with a skeleton-only variant at $12.09$
against the same $9.51$. The remaining appendices cover failed runs, additional RASR variants and
further views of the ground-truth and black-box tracks.

\section{Conclusion}\label{sec:conclusion}

\textbf{Limitations.} Several limitations remain in evaluation scope, method and measurement. Twelve black-box datasets leave track averages uncertain. The protocol fixes one expression per run while RASR returns a front, and the reported costs place GPU inference next to single-core CPU search. Constants at nearly every operator argument overparameterize refinement, and the bound of $2048$ rows and $11$ features costs accuracy on four black-box datasets. Finally, expression size, our only interpretability measure, cannot judge what an expression means to a domain expert.

\textbf{Discussion.} We see three implications. First, the prior is a primary design axis: at $24$M parameters, a model trained on an earlier-style prior ends at about twice our improvability (\appref{app:ablations}). We release ours as a modular basis for further work. Second, post-training specializes a base model. Needing no ground-truth expression, it could equally adapt to real data of a target domain or to other expression preferences. Third, our results are unlikely to mark the limit of the approach: gains from $24$M to $94$M parameters (\appref{app:posttraining-analysis}) are in line with early evidence of scaling laws~\citep{otte2025towards}, and larger training datasets may lift our bound on rows and
features.

Our results suggest that neural symbolic regression's deficit against search-based methods on real, noisy data stems from training rather than from the paradigm. With a synthetic prior built to resemble real data and reinforcement-learning post-training, RASR attains the best mean test $R^2$ rank on the black-box track of SRBench~2.0, with all three variants on the accuracy-complexity Pareto front, and the highest solution rate on Strogatz.
\subsection*{AI use statement}

We used LLM-based assistants to help implement the method and the training pipeline,
and we reviewed and tested all AI-assisted code. We also used them to polish the wording of text we wrote, to format \LaTeX{} tables, and to help
create figures. Some of the related work cited here was first suggested by an AI
assistant. We checked every cited work against the original publication. We did not
use generative AI to generate datasets, since our synthetic training data is sampled
from the prior described in \secref{sec:prior}. Nor did we use it to develop research
ideas, hypotheses, or theoretical frameworks, to design or give feedback on the
methodology or experiments, to choose experimental parameters, to formulate or prove mathematical claims, or to interpret results. Translation,
dataset cleaning, and qualitative data analysis do not apply to this work. We take full
responsibility for the content of this paper.

\subsection*{Reproducibility statement}

We aim for every result in this paper to be reproducible from the released artifacts and the
paper alone. Our source code is publicly available at \url{https://github.com/automl/rasr}. The weights of our models are hosted on Hugging Face at \url{https://hf.co/collections/davidotte/rasr}. In the paper itself, the synthetic prior is
described in \secref{sec:prior}, and \appref{app:prior} lists all of
its sampling hyperparameters, including our modifications to the TabICLv2 regression prior.
The architecture is given in \secref{sec:architecture} and \appref{app:architecture}.
\appref{app:training} lists the optimizer settings, schedules and compute budgets of both
training stages, and \appref{app:grpo} the post-training objective and its hyperparameters.
\appref{app:inference} specifies the sampling and constant-refinement settings, which are the
same in every experiment we report. We evaluate under the published protocols of SRBench~2.0
and the original SRBench, with their seeds and train/test splits (\secref{subsec:setup}).
\appref{app:protocol} states every choice that departs from these protocols or that they leave
open: the input-size caps and feature selection applied to RASR, median aggregation over
seeds, the improvability pools, and the configurations of Flash-ANSR and SNIP, the two
baselines we ran ourselves. \appref{app:failures} describes how failed runs are imputed. All
other baseline results are taken from the runs released with the respective benchmark.

\section*{Acknowledgements}

The authors acknowledge support by the state of Baden-Württemberg through bwHPC and the German Research Foundation (DFG) through grant INST 35/1597-1 FUGG and for funding under project number 455622343 (bwForCluster NEMO 2).
We further gratefully acknowledge the computing time made available on the high-performance computer at the NHR Center of TU Dresden. This center is jointly supported by the Federal Ministry of Research, Technology and Space of Germany and the state governments participating in the NHR (www.nhr-verein.de/unsere-partner).
This research was funded by the Deutsche Forschungsgemeinschaft (DFG, German Research Foundation) under grant number 417962828.
We acknowledge funding by the European Union (via ERC Consolidator Grant DeepLearning 2.0, grant no.~101045765). Views and opinions expressed are, however, those of the author(s) only and do not necessarily reflect those of the European Union or the European Research Council. Neither the European Union nor the granting authority can be held responsible for them. 

\begin{center}\includegraphics[width=0.3\textwidth]{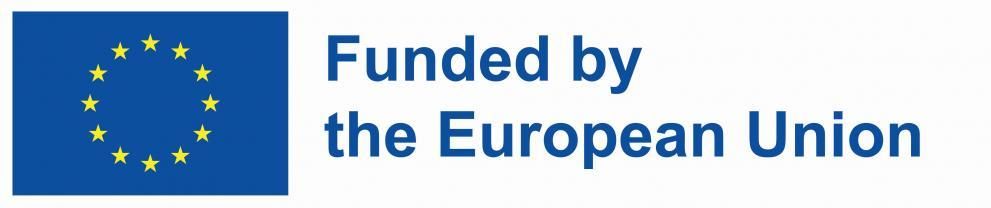}\end{center}

\newpage

\bibliography{bib/literature}
\bibliographystyle{iclr2027_conference}

\clearpage
\appendix

\let\appendixsection\section
\renewcommand{\section}{\FloatBarrier\appendixsection}

\section*{\centering Appendix}

\counterwithin{table}{section}
\counterwithin{figure}{section}

\section{Extended Related Work}\label{app:relatedwork}

This section expands \secref{sec:relatedwork} on neural symbolic regression, the family our
method belongs to, and closes with the reinforcement-learning line our post-training stage borrows
from. It is organized around the same recurring design choices, and states our own position on
each in place.

\paragraph{Synthetic training data.}

Every pre-trained method stands on a synthetic data generator.
For sampling the expressions, methods rely on the sampling strategy of
\citet{DBLP:conf/iclr/LampleC20},
which draws expression trees uniformly given their operator counts.
NeSymReS~\citep{DBLP:conf/icml/BiggioBNLP21} and Flash-ANSR~\citep{DBLP:journals/corr/abs-2602-08885} additionally
simplify the sampled expressions before training on them.
The generation of the accompanying datasets
has received far less attention. Inputs are typically sampled i.i.d.\ from simple
fixed distributions, a uniform one in NeSymReS and a mixture of Gaussians in
E2E~\citep{DBLP:conf/nips/KamiennydLC22}. Constants are sampled from fixed intervals chosen
independently of the data, with rejection sampling cleaning up the frequent
degenerate outcomes such as saturated exponentials and out-of-domain logarithms.
The resulting tables bear little resemblance to real
data, which is often noisy and contains correlated and redundant features. 
Since a pre-trained model can only be as good as its training data, we
regard this as a large untapped source of improvement in the recipe.

We therefore treat the data prior as a central design component rather than a fixed
preliminary, and construct inputs, constants, and targets accordingly
(\secref{sec:prior}).

\paragraph{Pre-training objectives.}

The standard recipe minimizes plain token-level cross-entropy against a single
reference expression, but a recent line of work argues for richer, multimodal
objectives. SNIP~\citep{DBLP:conf/iclr/MeidaniSRF24} uses a contrastive loss that aligns
numeric-data embeddings with embeddings of the corresponding expression, and the
model of \citet{DBLP:conf/iclr/LiLSWYLLT23} combines such a term with
the usual cross-entropy objective. Sym-Q~\citep{tian2025interactive} frames decoding as sequential
decision making, training with a contrastive loss plus an offline
reinforcement-learning objective.

We deliberately stay with plain cross-entropy, since these objectives mainly buy
searchability, a role filled in our pipeline by the post-training stage of
\secref{sec:training}.

\paragraph{Numerical constants.}

Existing methods split into two camps. Skeleton-based methods, among them
NeSymReS, Flash-ANSR, SNIP, Sym-Q, and the model
of \citet{DBLP:conf/iclr/LiLSWYLLT23}, predict only placeholder tokens and
delegate
all numeric values to a downstream optimizer. This keeps the task purely
symbolic, but it leaves the model no way to express a prior over plausible
constants, which could help avoid poor local minima and overfitting on
noisy data. End-to-end
methods, namely E2E and SymFormer~\citep{DBLP:journals/access/VastlKKDB24}, instead
aim to predict constants directly in high precision, with an optional refinement
step taking the predicted values as initialization. This places the full burden
of numeric precision on a discrete autoregressive decoder, a task at which
transformers are notoriously weak. A single mispredicted digit can leave the
refinement stage starting far from any good solution.

We take a middle position, emitting rounded constants (\secref{sec:prior}) that serve
only as an initialization region for the refinement of \secref{sec:inference}, so the
model steers the constant search without being trusted for precision.

\paragraph{Test-time compute and selection.}

Although not always framed as such, the methods above all spend test-time compute
on the same three ingredients: generating multiple candidate expressions, fitting
or refining each candidate's constants, and selecting a final one by training fit
and complexity. Some run these once in sequence, others interleave generation and
scoring. They differ in how many candidates they generate and in how they
generate them. Flash-ANSR
reports that recovery keeps improving as more expressions are sampled, whereas
NeSymReS and E2E settle for a small fixed number. On the generation mechanism,
NeSymReS, E2E, SymFormer, and Flash-ANSR use beam search or temperature sampling,
SNIP searches the latent space its pre-training constructed, and TPSR~\citep{DBLP:conf/nips/ShojaeeMFR23} wraps a Monte-Carlo tree search around a pre-trained decoder, using
fit-based rollout rewards to steer decoding.

We sample plainly with a small candidate budget (\secref{sec:inference}),
because post-training has already shifted the policy toward strong
candidates, and because an extensive search against a criterion computed on the observed
data invites overfitting when that data is small or noisy.

\paragraph{Deep symbolic regression.}

Reinforcement learning first entered symbolic regression as a per-dataset search.
Deep symbolic regression (DSR)~\citep{DBLP:conf/iclr/PetersenLMSKK21} trains a policy network from
scratch on each problem, optimizing a risk-seeking policy gradient that rewards
the best sample in a batch rather than the average one, which reflects that only
the single best expression is ultimately returned. We build on the same idea, but
apply it a single time, as a post-training stage on our general pre-trained
model, so that the search cost is amortized across all future datasets instead of
being repaid on each one. Our objective keeps the risk-seeking motivation of DSR
and realizes it within a group-relative policy optimization scheme
(\appref{app:grpo}), which matches the best-of-$n$ metric that our
sampling-based inference actually targets.

\section{Synthetic Prior Details}\label{app:prior}

This section specifies the pre-training prior of \secref{sec:prior} in full, following the
order of the pipeline in \figref{fig:prior}. All values correspond to the configuration used
during training.

\subsection{Skeleton Sampling}\label{app:skeleton}

\tabref{tab:app-skeleton} lists the structural ranges of the skeleton sampler and the
categorical operator weights, which are normalized separately within the binary and unary
sets. Given the sampled number of binary operators, the binary tree is drawn uniformly over all
trees of that size with the counting procedure of \citet{DBLP:conf/iclr/LampleC20}. Its leaves
receive the $n_v$ distinct variables (or as many as the tree has leaves), and the remaining
leaves are filled with variables drawn with replacement from these, before all leaves are
shuffled. The unary operators are then inserted at positions drawn uniformly with replacement
from the prefix sequence, each wrapping the subtree rooted at its position.

\begin{table}[tbp]
  \centering
  \caption{\textbf{Skeleton-sampling hyperparameters.} The upper bound on the
  binary-operator count grows with the number of distinct variables $n_v$, so
  that expressions over more variables may be structurally larger.}
  \label{tab:app-skeleton}
  \begin{tabular}{lc}
    \hline
    Parameter & Value \\
    \hline
    Distinct variables $n_v$            & uniform $\{1,\dots,11\}$ \\
    Binary ops $n_b$                    & uniform $\{0,\dots,8 + n_v\}$ \\
    Unary ops $n_u$                     & uniform $\{0,\dots,5\}$ \\
    Max expression length               & $512$ tokens \\
    Constant mantissa                   & $3$ significant figures \\
    Constant magnitude bound            & $[10^{-4}, 10^{4}]$ \\
    \hline
    \multicolumn{2}{c}{Binary-operator weights} \\
    \hline
    $+$ : $10$ \quad\quad $\times$ : $5$ \quad\quad $\div$ : $5$  &  \\
    \hline
    \multicolumn{2}{c}{Unary-operator weights} \\
    \hline
    $x^2,x^3$ & $1$ \\
    $x^4,x^5$ & $0.5$ \\
    $x^{1/2},x^{1/3}$ & $1$ \\
    $x^{1/4},x^{1/5}$ & $0.5$ \\
    $\mathrm{abs},\mathrm{inv},\sin,\cos,\tanh,\exp$ & $1$ \\
    $\log$ & $0.5$ \\
    \hline
  \end{tabular}
\end{table}

\paragraph{Constant placeholders.}
Placeholders are inserted in a single recursive pass over the tree with the following rules.
The argument of every unary operator is wrapped affinely, $g(t) \rightarrow g(c_0 t + c_1)$.
Every summand of a sum that is not itself a sum receives a multiplicative scale, and only the
outermost sum of a chain receives an additive bias. A product chain receives a single scale
and bias at its outermost node, $c_0 (a \cdot b) + c_1$. A quotient scales its numerator and
shifts its denominator, $c_0 a / (b + c_1) + c_2$. Variable leaves receive a scale $c\,x_i$
unless their parent already supplies one. These rules skip exactly the constants that would be
redundant: a scale on each factor of a product collapses under the product's own scale, as in
$c_0(c_1x_1)(c_2x_2) \equiv c\,x_1x_2$, and a scale inside a sum that is itself scaled would
likewise collapse. The global scale and offset of the expression are not emitted, as they are
absorbed by the target standardization (\appref{app:standardization}).

\paragraph{Simplification and filtering.}
The skeleton with placeholders is then simplified with SimpliPy~\citep{DBLP:journals/corr/abs-2602-08885},
which collapses remaining redundant terms, including constant placeholders that merge. The
result is rejected if it contains non-finite tokens, contains no variable, or exceeds the
maximum length. Finally, the variables that survive simplification are re-indexed to a
contiguous prefix $x_1,\dots,x_{n_v}$, which fixes their correspondence to the first $n_v$
input columns (\appref{app:inputs}).

\subsection{Input Sampling and TabICLv2 Modifications}\label{app:inputs}\label{app:tabicl}

Prior neural symbolic regression methods draw $X$ from simple fixed distributions, such as a
uniform distribution~\citep{DBLP:conf/icml/BiggioBNLP21} or a mixture of
Gaussians~\citep{DBLP:conf/nips/KamiennydLC22}. We instead use the TabICLv2 regression prior
with numerical features only, from which we draw a pool of twelve columns, each standardized to
zero mean and unit variance. The first $n_v$ columns become the variables and the last one the
noise node (\appref{app:dataset}). The number of distractor columns is drawn from a half-normal
with scale $1.5$, rounded, and capped by the number of remaining columns. After the target has
been computed, all observed feature columns, variables and distractors alike, are randomly
permuted and the variables of the expression renamed accordingly, so that a column's position
carries no information about whether the expression uses it.

We use the TabICLv2 prior essentially as published, with the targeted changes summarized in
\tabref{tab:app-tabicl}, all of which serve to make its columns suitable as the continuous
support of a symbolic target. First, we bias the mixture of feature-generating function types
towards smooth ones, giving equal weight to multilayer-perceptron, linear, quadratic,
Gaussian-process, and product functions while down-weighting the piecewise-constant tree and
discretization functions to a single component each. Second, for the same reason, we remove
the discontinuous, step-like activations from the prior's activation pool. Third, where
discrete functions do occur, we raise their minimum granularity, so that the resulting columns
still have enough distinct levels to act as continuous regressors. Finally, we add a
per-column diversity filter that rejects a sampled dataset if any feature column has fewer than
eight distinct value clusters or a single value cluster covering more than a third of the rows.
Together these changes steer the prior away from degenerate, low-cardinality inputs while
leaving the bulk of it untouched.

\begin{table}[tbp]
  \centering
  \caption{\textbf{Changes to the published TabICLv2 regression prior.} Arrows denote a
  changed default value.}
  \label{tab:app-tabicl}
  \begin{tabular}{ll}
    \hline
    Component & Setting \\
    \hline
    Function-type mixture        & $4\times$\{mlp, lin, quad, gp, prod\}, $1\times$\{tree, disc\} \\
    Removed activations          & sign, Heaviside, unit-interval, argmax indicators \\
    Min.\ random-tree depth      & $1 \rightarrow 3$ \\
    Min.\ discretization centers & $2 \rightarrow 5$ \\
    Per-column diversity filter  & added ($\geq 8$ clusters; no cluster $> 1/3$ of rows) \\
    \hline
  \end{tabular}
\end{table}

\subsection{Bottom-Up Constant Sampling}\label{app:constants}

Constants are filled in a single bottom-up pass in which every subtree is evaluated on $X$
before the constants of its parent are drawn. For the affine wrapper $g(c_0 t + c_1)$ of a
unary operator, the scale is drawn log-normally, $\log c_0 \sim \mathcal{N}(\mu, \sigma^2)$,
divided by the empirical standard deviation of the child values $t$, and given a random sign
except for the positive-domain operators. The shift $c_1$ is sampled in an operator-specific,
domain-aware way, with four shift types listed in \tabref{tab:app-constants}:
\begin{itemize}
  \item \emph{Generic:} $c_1 = -c_0 \bar{t} + \delta$ with $\delta \sim \mathcal{N}(m, \sigma_1^2)$,
    so that the mean of the argument is centered at $m$ with spread $\sigma_1$.
  \item \emph{Positive} ($\log$, even roots): a distance $d$ with
    $\log d \sim \mathcal{N}(\log m, \sigma_1^2)$ is drawn and $c_1 = d - \min(c_0 t)$, so the
    minimum of the argument equals $d > 0$ by construction.
  \item \emph{Signed} ($\mathrm{inv}$ and the denominator of $\div$): with probability $0.4$
    each the argument is placed entirely on the positive or the negative side, at a distance $d$
    from zero with $\log d \sim \mathcal{N}(-0.69, 0.5^2)$, and with probability $0.2$ it is
    centered, $c_1 = -\overline{c_0 t} + \mathcal{N}(0, 0.5^2)$, so that it may cross zero. A
    denominator that simplified to a bare constant only uses the two one-sided branches.
  \item \emph{Straddling} ($\mathrm{abs}$): $c_1$ is drawn from $\mathcal{N}(-\overline{c_0 t}, \sigma_1^2)$
    truncated such that the argument takes both signs and each of its extremes lies at least
    $5\%$ of its range away from zero, which ensures that $\mathrm{abs}$ actually bends.
\end{itemize}

\begin{table}[tbp]
  \centering
  \caption{\textbf{Per-operator affine-constant distributions.} $c_0$ is log-normal with
  the listed $(\mu,\sigma)$ on the log scale (then normalized by the child's
  standard deviation); $c_1$ follows the listed shift type with parameters
  $(m,\sigma_1)$. Powers, roots, and the two trigonometric functions share
  parameters and are grouped. The signed shift of $\mathrm{inv}$ is parameterized in the text.}
  \label{tab:app-constants}
  \small
  \setlength{\tabcolsep}{5pt}
  \begin{tabular}{lccl}
    \hline
    Operator(s) & $c_0$ $(\mu,\sigma)$ & $c_1$ params & Shift type \\
    \hline
    $\sin, \cos$                       & $(0.45, 1.0)$  & $(0, 2.0)$   & generic \\
    $\exp$                             & $(-0.69, 0.8)$ & $(0, 1.0)$   & generic \\
    $\tanh$                            & $(0.0, 0.7)$   & $(0, 1.5)$   & generic \\
    $x^2,x^3,x^4,x^5,x^{1/3},x^{1/5}$  & $(0.45, 1.0)$  & $(0, 4.0)$   & generic \\
    $\log$                             & $(0.4, 0.8)$   & $(1.5, 1.0)$ & positive \\
    $x^{1/2}, x^{1/4}$                 & $(1.0, 0.8)$   & $(1.5, 1.0)$ & positive \\
    $\mathrm{inv}$                     & $(-0.69, 0.5)$ & see text     & signed \\
    $\mathrm{abs}$                     & $(0.0, 0.5)$   & $(0, 1.5)$   & straddling \\
    \hline
  \end{tabular}
\end{table}

\paragraph{Sums.}
The scales of a sum are set jointly rather than per term. All scaled terms of an additive
chain are evaluated first, and each scale is set to
$c_i = \pm\, \sigma_{\mathrm{base}} / \sigma_i \cdot e^{\xi_i}$ with a random sign,
$\xi_i \sim \mathcal{N}(0, 0.25^2)$, $\sigma_i$ the empirical standard deviation of term $i$,
and $\sigma_{\mathrm{base}}$ the geometric mean of all $\sigma_i$. If the chain has only one
scaled term, $\sigma_{\mathrm{base}}$ is the spread of the remaining unscaled part instead.
Every term thus contributes a comparable spread, so that none vanishes or dominates. When the
argument of a unary operator is itself a two-term sum, its overall spread is drawn from that
operator's $c_0$ distribution and split between the two terms with a log-ratio
$\mathcal{N}(0, 0.7^2)$. All remaining constants, such as the biases of binary operators and
scales outside of sums, are drawn from $\mathcal{N}(0, 5^2)$.

\paragraph{Validation and filtering.}
Inputs are drawn with $25\%$ more rows than the target number of observations $n$, so that
filtering can remove rows without shrinking the dataset. A pair is rejected if any raw
constant violates $10^{-4} < \lvert c \rvert < 10^{4}$. After noise has been added
(\appref{app:dataset}), rows with a non-finite value are removed, and if more than the surplus
are affected, the pair is rejected. Gross target outliers are then removed with an
interquartile-range filter, discarding points outside
$[\,Q_1 - 5\,\mathrm{IQR},\; Q_3 + 5\,\mathrm{IQR}\,]$, and the dataset is truncated to $n$
rows, being rejected if fewer remain. The resulting target must then be well-behaved: its
variance must exceed $10^{-6}$, its maximum absolute value must stay below $10^{6}$, and the
largest $1\%$ of its squared deviations from the mean must contribute less than $90\%$ of the
sum of all squared deviations (a heavy-tail guard). Finally, the dataset is discarded if its empirical target
standard deviation falls outside $[10^{-6}, 10^{3}]$.

\subsection{Noise and Standardization}\label{app:dataset}\label{app:standardization}

\tabref{tab:app-dataset} lists the dataset- and noise-level settings. With probability $0.4$ a
dataset stays clean. Otherwise, a strength $s$ is drawn from an exponential distribution with
mean $0.08$, and each target is perturbed as
\begin{equation}
  \tilde{y}_i \;=\; y_i + s\, \sigma_y\, \sigma_i\, z_i,
  \label{eq:app-noise}
\end{equation}
where $\sigma_y$ is the standard deviation of the clean target, $z$ a noise shape with zero mean
and unit variance over the dataset, and $\sigma_i$ a per-row scale with mean one. Let $\nu$ denote
the noise node. The two noise forms, drawn with equal probability, differ in $z$ and $\sigma_i$.

\textbf{Gaussian form.} Here $z_i \sim \mathcal{N}(0, 1)$, empirically standardized, and the
per-row scale is a blend between a constant and a softplus profile of the noise node,
\begin{equation}
  \sigma_i \;\propto\; \min\!\Big((1 - w) + w\, \frac{\mathrm{softplus}(\nu_i)}{\overline{\mathrm{softplus}(\nu)}},\; 3\Big),
  \qquad w \sim \mathcal{U}[0, 1],
\end{equation}
normalized to mean one and clipped again at $3$. The blend weight $w$ interpolates between
homoscedastic ($w = 0$) and fully input-correlated heteroscedastic noise ($w = 1$), and the
clipping bounds the per-row scale at three times its mean. Since softplus is positive and
monotone, the noise level is correlated with whatever the noise node is correlated with in the
sampled SCM.

\textbf{Structured form.} Here $\sigma_i = 1$ and $z$ is the noise node passed through a random
activation and then standardized to zero mean and unit variance. The activation is sampled as
the composition: standardize to unit variance, apply a random positive rescaling, an optional
sign flip of the input, one activation drawn uniformly from \{identity, LeakyReLU, ELU, SELU,
SiLU, ReLU, ReLU6, softplus, sigmoid, $\exp$, square, $\lvert\cdot\rvert$,
$\sqrt{\lvert\cdot\rvert}$\}, and an optional sign flip of the output.

\textbf{Standardization.} The target is standardized with the empirical mean $\mu_y$ and
standard deviation $\sigma_y$ of the final, possibly noisy target, so that the label is defined
with respect to the data the model actually observes. The expression is rewritten as
$(f - \mu_y)/\sigma_y$, where $\mu_y$ is folded into the top-level additive constant of $f$
whenever one exists, which avoids a redundant subtraction.

\begin{table}[tbp]
  \centering
  \caption{\textbf{Dataset- and noise-level hyperparameters.}}
  \label{tab:app-dataset}
  \small
  \begin{tabular}{lc}
    \hline
    Parameter & Value \\
    \hline
    Observations $n$                 & uniform integer $[8, 1024]$ \\
    Distractor columns               & half-normal, scale $1.5$, capped by pool size \\
    \hline
    Clean-dataset fraction           & $0.4$ \\
    Noise strength $s$               & $\mathrm{Exponential}(\text{mean } 0.08)$ \\
    Gaussian / structured form       & $0.5$ / $0.5$ \\
    Blend weight $w$ (Gaussian form) & $\mathcal{U}[0, 1]$ \\
    Max.\ per-row scale $\sigma_i$ (Gaussian form) & $3$ \\
    \hline
  \end{tabular}
\end{table}

\subsection{Tokenization}\label{app:tokenization}

Constants are rounded to three significant figures only after the standardization rewrite, so
the rounding applies to the values the model is trained to predict. Each constant is then
written in scientific notation and emitted digit by digit. For example, $-0.0421 = -4.21 \times
10^{-2}$ becomes \texttt{neg 4 2 1 e- 2}: a negative constant is prefixed with the ordinary
unary negation operator \texttt{neg}, followed by the mantissa digits without the decimal point
and with trailing zeros dropped, the exponent sign \texttt{e+} or \texttt{e-}, and a single
exponent digit. Numeric values of any magnitude thus share a fixed alphabet of twelve tokens,
the ten digits and the two exponent signs, rather than requiring one token per value.

\subsection{Prior Statistics}

\begin{figure}[tbp]
  \centering
  \includegraphics[width=\linewidth]{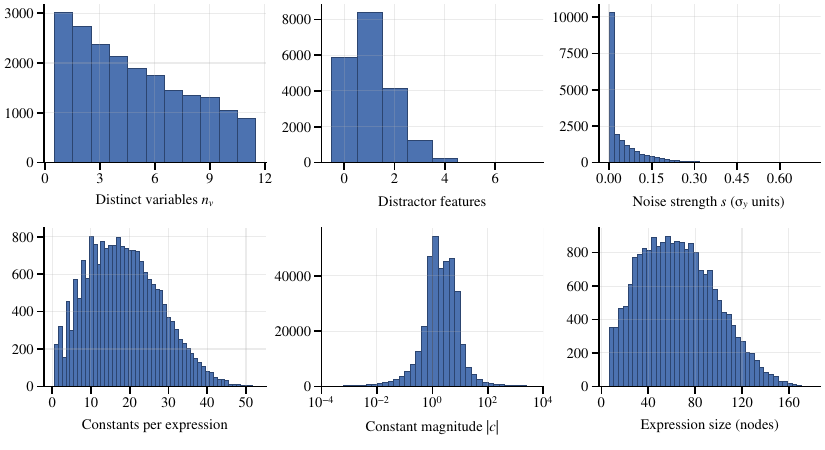}
  \caption{\textbf{\boldmath Statistics of the pre-training prior over $20{,}000$ sampled pairs.} Top
  row, left to right: the number of distinct variables $n_v$ in the expression, the number of
  distractor features appended to the dataset, and the noise strength $s$ (in units of the clean
  target's standard deviation $\sigma_y$). Bottom row, left to right: the number of constants per
  expression, the magnitude of the sampled constants (log scale, bounded by the sampler's own
  rejection threshold), and the expression size in nodes.}
  \label{fig:prior-statistics}
\end{figure}

To characterize what the pipeline actually produces, \figref{fig:prior-statistics} shows the
marginal distributions of its main quantities over a large batch of sampled pairs. The number
of distinct variables is close to uniform by construction, while the number of distractor
features is concentrated at small values with a long tail. Most datasets carry little or no
noise, consistent with the $40\%$ clean fraction and the exponential's mean of $0.08$, but the
distribution has a long tail toward heavily corrupted targets. Expressions typically carry
several constants, and the bottom-up sampler keeps their magnitude concentrated within a few
orders of magnitude of one rather than filling the full range the sampler permits, consistent
with the intent of conditioning on the data. Expression size is spread over a wide range,
giving the model exposure to both small and considerably larger expressions.

\section{Architecture}\label{app:architecture}

\begin{figure}[tbp]
  \centering
  \resizebox{0.77\linewidth}{!}{
{%
\definecolor{archmuted}{HTML}{78848F}
\definecolor{archattn}{HTML}{E7DCF6}
\definecolor{archattnline}{HTML}{9B7FD4}
\definecolor{archff}{HTML}{D6E6F7}
\definecolor{archffline}{HTML}{6E9FD4}
\definecolor{archembed}{HTML}{D9EEDF}
\definecolor{archembedline}{HTML}{6DB585}
\definecolor{archdata}{HTML}{FBE7CD}
\definecolor{archdataline}{HTML}{DDA45C}
\definecolor{archnorm}{HTML}{ECEFF2}
\definecolor{archnormline}{HTML}{BFC7CF}
\rmfamily
\begin{tikzpicture}[
    every node/.style={text=black},
    unit/.style={
      rounded corners=2.5pt, draw, line width=0.7pt, align=center,
      font=\small, minimum height=7mm, text width=32mm, inner sep=3pt,
    },
    attn/.style={unit, fill=archattn, draw=archattnline},
    ff/.style={unit, fill=archff, draw=archffline},
    embed/.style={unit, fill=archembed, draw=archembedline},
    data/.style={unit, fill=archdata, draw=archdataline},
    norm/.style={
      rounded corners=1.5pt, draw=archnormline, fill=archnorm, line width=0.5pt,
      font=\small, text=black, minimum height=3.4mm, text width=32mm,
      align=center, inner sep=1pt,
    },
    stack/.style={
      draw=archmuted!60, line width=0.6pt, dashed, rounded corners=3pt,
      inner sep=4pt,
    },
    plus/.style={
      circle, draw=archmuted, line width=0.6pt, fill=white, inner sep=0pt,
      minimum size=3.2mm, font=\small,
    },
    flow/.style={-{Latex[length=4pt,width=3.4pt]}, draw=archmuted,
                 line width=0.8pt},
    skip/.style={draw=archmuted!70, line width=0.6pt, rounded corners=2pt},
    cellon/.style={fill=archattnline!35, draw=archattnline!60, line width=0.3pt},
    celloff/.style={fill=white, draw=archnormline, line width=0.3pt},
    datacell/.style={fill=white, draw=archdataline!70, line width=0.35pt,
                     minimum width=3.2mm, minimum height=3.2mm, inner sep=0pt},
    targetcell/.style={datacell, fill=archdataline!45},
    latcell/.style={datacell, draw=archembedline!70},
    latqueried/.style={latcell, fill=archembedline!45},
    lane/.style={font=\small\bfseries, text=black},
  ]

  \node[data, draw=none, fill=none, minimum height=21mm] (tbl) at (0,0) {};
  \node[font=\small, anchor=north] (tbllab) at ([yshift=-3pt]tbl.north) {Input Table};
  \begin{scope}[shift={($(tbl.south)+(0,6.6mm)$)}]
    \foreach \c in {0,1,2} {
      \foreach \r in {0,1,2} {
        \node[datacell] at (0.36*\c-0.36, 0.36*\r) {};
      }
    }
    \foreach \r in {0,1,2} { \node[targetcell] at (-0.36, 0.36*\r) {}; }
    \node[font=\small, text=black, anchor=north] (tblcap) at (0, -0.28)
      {Target column and $d$ features};
  \end{scope}
  \node[embed, above=6mm of tbl] (cellenc)
    {Cell Encoding\\[-1pt]\small \texttt{float32} bits $\to$ linear};
  \node[plus, above=4mm of cellenc] (pe) {$+$};
  \node[font=\small, text=black, right=1.5mm of pe, align=left]
    {Column position};

  \node[norm, above=5mm of pe] (n1) {RMSNorm};
  \node[attn, above=1.2mm of n1] (satt)
    {Feature-Axis Attention\\[-1pt]\small QK-Norm, Gated};
  \node[plus, above=1.2mm of satt] (p1) {$+$};

  \node[norm, above=1.2mm of p1] (n2) {RMSNorm};
  \node[attn, above=1.2mm of n2] (fatt)
    {Sample-Axis Attention\\[-1pt]\small QK-Norm, QASSMax, Gated};
  \node[plus, above=1.2mm of fatt] (p2) {$+$};

  \node[norm, above=1.2mm of p2] (n3) {RMSNorm};
  \node[ff, above=1.2mm of n3] (eff) {Feed-Forward};
  \node[plus, above=1.2mm of eff] (p3) {$+$};

  \node[stack, inner sep=7pt, fit=(n1)(satt)(n2)(fatt)(n3)(eff)(p3)]
    (encstack) {};
  \node[lane, fill=white, inner xsep=2pt, inner ysep=0pt, anchor=west]
    at ([xshift=4mm]encstack.north west) {$L\times$ Encoder block};

  \node[data, draw=none, fill=none, above=6mm of encstack,
    minimum height=21mm] (lat) {};
  \node[font=\small, anchor=north] (latlab) at ([yshift=-3pt]lat.north)
    {Per-Cell Latents};
  \begin{scope}[shift={($(lat.south)+(0,6.6mm)$)}]
    \foreach \c in {0,1,2} {
      \foreach \r in {0,1,2} {
        \node[latcell] at (0.36*\c-0.36, 0.36*\r) {};
      }
    }
    \foreach \r in {0,1,2} { \node[latqueried] at (-0.36, 0.36*\r) {}; }
    \node[font=\small, text=black, anchor=north] (latcap) at (0, -0.28)
      {Shaded cells feed the decoder};
    \coordinate (latgride) at (0.56, 0.36);
  \end{scope}

  \node[data] (toks) at (6.6,0)
    {Expression Tokens\\[-1pt]\small Prefix sequence};
  \node[embed, above=6mm of toks] (temb)
    {Token Embedding\\[-1pt]\small Tied to the output projection};
  \node[plus, above=4mm of temb] (tpe) {$+$};
  \node[font=\small, text=black, right=1.5mm of tpe, align=left]
    {Token position};

  \node[norm, above=5mm of tpe] (m1) {RMSNorm};
  \node[attn, above=1.2mm of m1] (xatt)
    {Cross-Attention\\[-1pt]\small QK-Norm, QASSMax, Gated};
  \node[plus, above=1.2mm of xatt] (q1) {$+$};

  \node[norm, above=1.2mm of q1] (m2) {RMSNorm};
  \node[attn, above=1.2mm of m2] (catt)
    {Masked Self-Attention\\[-1pt]\small QK-Norm, RoPE, Gated};
  \node[plus, above=1.2mm of catt] (q2) {$+$};

  \node[norm, above=1.2mm of q2] (m3) {RMSNorm};
  \node[ff, above=1.2mm of m3] (dff) {Feed-Forward};
  \node[plus, above=1.2mm of dff] (q3) {$+$};

  \node[stack, inner sep=7pt, fit=(m1)(xatt)(m2)(catt)(m3)(dff)(q3)]
    (decstack) {};
  \node[lane, fill=white, inner xsep=2pt, inner ysep=0pt, anchor=west]
    at ([xshift=4mm]decstack.north west) {$L\times$ Decoder block};

  \node[embed, above=6mm of decstack] (head)
    {Output Projection\\[-1pt]\small Tied embedding, softmax};
  \node[data, above=5mm of head] (out) {Next-Token Distribution};

  \draw[flow] (tbllab.north) -- (cellenc);
  \draw[flow] (cellenc) -- (pe);
  \draw[flow] (pe) -- (n1);
  \draw[flow] (n1) -- (satt);
  \draw[flow] (satt) -- (p1);
  \draw[flow] (p1) -- (n2);
  \draw[flow] (n2) -- (fatt);
  \draw[flow] (fatt) -- (p2);
  \draw[flow] (p2) -- (n3);
  \draw[flow] (n3) -- (eff);
  \draw[flow] (eff) -- (p3);
  \draw[flow] (p3) -- (latcap.south);

  \draw[flow] (toks) -- (temb);
  \draw[flow] (temb) -- (tpe);
  \draw[flow] (tpe) -- (m1);
  \draw[flow] (m1) -- (xatt);
  \draw[flow] (xatt) -- (q1);
  \draw[flow] (q1) -- (m2);
  \draw[flow] (m2) -- (catt);
  \draw[flow] (catt) -- (q2);
  \draw[flow] (q2) -- (m3);
  \draw[flow] (m3) -- (dff);
  \draw[flow] (dff) -- (q3);
  \draw[flow] (q3) -- (head);
  \draw[flow] (head) -- (out);

  \foreach \a/\b in {n1/p1, n2/p2, n3/p3} {
    \draw[skip] ([yshift=-1mm]\a.west) -- ++(-2.5mm,0)
      |- ([xshift=-2.5mm]\b.west) -- (\b.west);
  }
  \foreach \a/\b in {m1/q1, m2/q2, m3/q3} {
    \draw[skip] ([yshift=-1mm]\a.east) -- ++(2.5mm,0)
      |- ([xshift=2.5mm]\b.east) -- (\b.east);
  }

  \draw[flow] (latgride) -- ++(31mm,0) |- (xatt.west);

  \begin{scope}[shift={($(satt.east)+(0.45cm,-0.26cm)$)}]
    \foreach \c in {0,1,2} {
      \foreach \r in {0,1,2} {
        \node[celloff, minimum width=2.2mm, minimum height=2.2mm, inner sep=0pt]
          at (0.26*\c, 0.26*\r) {};
      }
    }
    \foreach \c in {0,1,2} {
      \node[cellon, minimum width=2.2mm, minimum height=2.2mm, inner sep=0pt]
        at (0.26*\c, 0.26) {};
    }
  \end{scope}

  \begin{scope}[shift={($(fatt.east)+(0.45cm,-0.26cm)$)}]
    \foreach \c in {0,1,2} {
      \foreach \r in {0,1,2} {
        \node[celloff, minimum width=2.2mm, minimum height=2.2mm, inner sep=0pt]
          at (0.26*\c, 0.26*\r) {};
      }
    }
    \foreach \r in {0,1,2} {
      \node[cellon, minimum width=2.2mm, minimum height=2.2mm, inner sep=0pt]
        at (0.26, 0.26*\r) {};
    }
  \end{scope}
\end{tikzpicture}%
}}
  \caption{\textbf{Overview of the model architecture.} The table, whose shaded first column holds
  the target, is encoded cell by cell and refined by a stack of encoder blocks that
  alternate attention over its two axes. The grids beside them mark the cells attended
  together, along a row for feature-axis attention and down a column for sample-axis
  attention. The encoder returns one latent per cell, and the decoder cross-attends to
  those of the target column alone, shaded in the grid, while generating the expression as
  a prefix token sequence. Every sublayer is pre-normalized and wrapped in a residual
  connection.}
  \label{fig:architecture}
\end{figure}
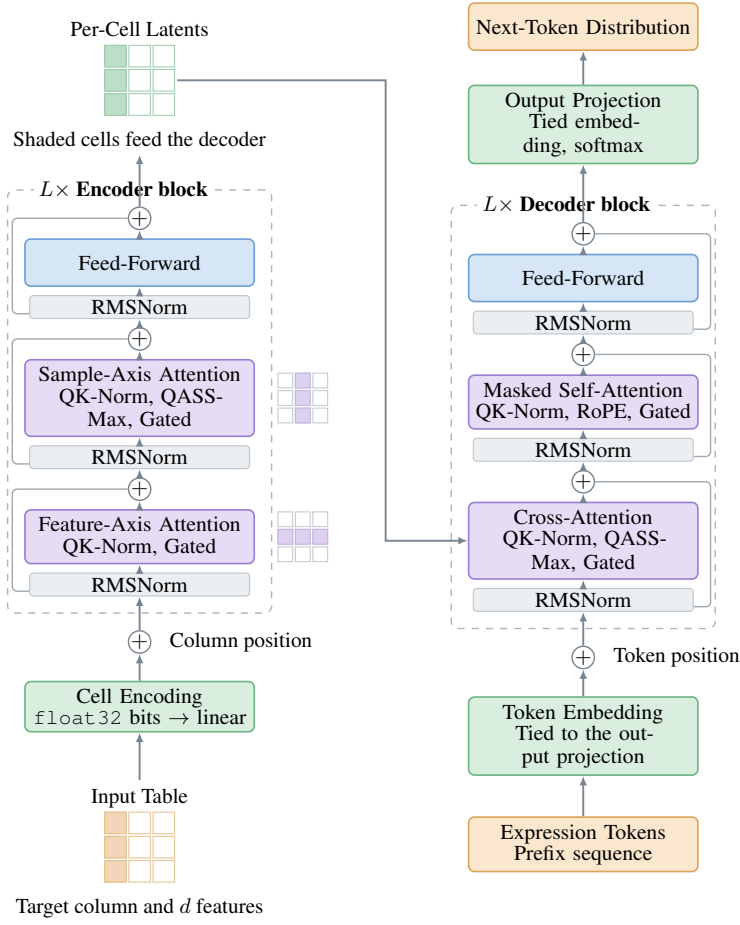

This section expands \secref{sec:architecture} and specifies every component of the model.
\figref{fig:architecture} gives an overview, and \tabref{tab:model-sizes} the two configurations
we train.

\paragraph{Table input and cell encoding.} The input is a table of shape $(n, d{+}1)$
whose first column holds the target and whose remaining columns hold the $d$ features.
Within a batch, tables are padded to a common shape and a boolean mask records the valid
entries. Every scalar cell is encoded through its raw IEEE-754 representation: the $32$
bits of the \texttt{float32} value are mapped to a $\{-1,+1\}^{32}$ vector, which a linear
layer projects to the model dimension~\citep{DBLP:conf/icml/BiggioBNLP21}. This gives the
model direct access to the magnitude and sign of every value. A standard sinusoidal
positional embedding~\citep{DBLP:conf/nips/VaswaniSPUJGKP17} over the column index is added
so that the encoder can distinguish the target column from the features and the features
from one another.

\paragraph{Alternating axial-attention encoder.} Following the encoder design of
TabPFNv2~\citep{DBLP:journals/nature/HollmannMPKKHSH25}, each block applies attention along
the feature axis, letting every cell attend to the other columns in its row, and along the
sample axis, letting every cell attend to the same column in other rows, followed by a
single position-wise feed-forward network. This factorization avoids full attention over
all $n(d{+}1)$ cells while keeping the model expressive. Sample-axis attention treats the
rows as an unordered set, so the encoder is permutation-equivariant over samples and
accepts a variable number of rows $n$. The column positional embedding breaks this symmetry
along the feature axis. After the final block, the representation at the target column of
each row is taken as that row's summary, yielding the $n$ per-sample latents the decoder
attends to.

\paragraph{Decoder.} The decoder is a stack of standard transformer decoder blocks. Each
block first cross-attends to the $n$ per-sample latents emitted by the encoder, then
applies causal self-attention with rotary position embeddings
(RoPE)~\citep{DBLP:journals/corr/abs-2104-09864} and finally a position-wise feed-forward
network. Token embeddings use a learned table with additive sinusoidal absolute positions,
and the embedding matrix is tied to the output
projection~\citep{DBLP:conf/eacl/PressW17}.

\paragraph{Length-robust attention (QASSMax).} Standard attention scales the query-key dot
product by a fixed factor $1/\sqrt{d_h}$, which is independent of the number of keys. In
our encoder the relevant context is the set of samples, whose size $n$ varies by two orders
of magnitude across datasets, and with a fixed scale the softmax over $n$ logits flattens
as $n$ grows. Scalable-softmax (SSMax)~\citep{DBLP:journals/corr/abs-2501-19399} counteracts
this by scaling the logits with the context length. We use the query-adaptive generalization
from TabICLv2~\citep{DBLP:journals/corr/abs-2602-11139}, QASSMax, which rescales each query
vector $q$ by a factor depending on both $n$ and $q$ itself,
\begin{equation}
  \tilde{q} \;=\; q \,\odot\, \big[\, s(\log n) \,\odot\, \big(1 + \tanh(m(q))\big) \,\big],
  \label{eq:qassmax}
\end{equation}
where $\odot$ is elementwise multiplication, $s(\cdot)$ is a small MLP mapping $\log n$ to a
per-head, per-dimension base scale, and $m(\cdot)$ is a second small MLP producing a
per-query modulation mapped into $(0, 2)$. We apply QASSMax only along the sample axis. As
the feature count $d$ stays small, the feature axis keeps the ordinary scale.

\paragraph{Normalization and gating.} All blocks use pre-normalization with
RMSNorm~\citep{DBLP:conf/nips/ZhangS19a}, and we additionally normalize the queries and keys
before their dot product (QK-norm), which bounds the attention logits and has been reported
to stabilize training, particularly in low precision~\citep{DBLP:conf/emnlp/HenryDPC20}.
Feed-forward networks use a SiLU activation~\citep{DBLP:journals/nn/ElfwingUD18}, and
residual additions are accumulated in \texttt{float32}. Following recent work on gated
attention~\citep{DBLP:conf/nips/QiuWZHWYMYHHLZL25}, each attention head is further modulated
by an input-dependent scalar gate: from the same pre-normalized input $x$ that feeds the
query, key and value projections, a linear map followed by a sigmoid produces one gate per
head, $g = \sigma(W_g x) \in (0,1)^{H}$, and the output of head $h$ is multiplied by $g_h$
before the heads are concatenated and projected. The gating lets the model suppress
uninformative heads.

\begin{table}[tbp]
  \centering
  \caption{\textbf{Configuration of the two model sizes.}}
  \label{tab:model-sizes}
  \begin{tabular}{lccccc}
    \hline
    Model & Params & Model dim. & Enc.\ layers & Dec.\ layers & Heads \\
    \hline
    $24$M & ${\sim}24.37$M & $384$ & $5$  & $5$  & $6$ \\
    $94$M & ${\sim}94.13$M & $512$ & $11$ & $11$ & $8$ \\
    \hline
  \end{tabular}
\end{table}

\section{Training Details}\label{app:training}

Pre-training uses AdamW~\citep{DBLP:conf/iclr/LoshchilovH19} ($\beta_1 = 0.9$, $\beta_2 = 0.98$, weight
decay $0.1$) with a schedule that warms the learning rate up linearly over the first $5\%$
of steps and then decays it with a cosine profile~\citep{DBLP:conf/iclr/LoshchilovH17} to
$10\%$ of its peak. Gradients are clipped to a global norm of $1.0$, and training runs in
\texttt{bfloat16} mixed precision with \texttt{float32} reductions. The peak learning rate
is set per model size and was selected with a coarse logarithmic grid search; the batch
size and remaining optimization hyperparameters follow standard practice and a small number
of manual trials. \tabref{tab:pre-training-budget} lists the per-size budgets.

\begin{table}[htbp]
  \centering
  \caption{\textbf{Pre-training budget per model size.}}
  \label{tab:pre-training-budget}
  \small
  \begin{tabular}{lcccccc}
    \hline
    Model & Batch Size & Steps & Datasets & Peak LR & GPUs & Duration \\
    \hline
    $24$M & $64$  & $250$k & ${\approx}16$M & $4.6{\times}10^{-4}$ & $4\times$ L40S & ${\approx}31$\,h \\
    $94$M & $256$ & $242$k & ${\approx}62$M & $2.2{\times}10^{-4}$ & $8\times$ H200 & ${\approx}71$\,h  \\
    \hline
  \end{tabular}
\end{table}

Post-training reuses the AdamW settings of pre-training under a cosine schedule with a $5\%$
warmup and the peak learning rate set to $10\%$ of the pre-training learning rate,
corresponding to the annealed rate at the end of pre-training. We post-train for $2{,}000$ steps
with a global batch of $32$ datasets, ${\approx}64$k datasets in total, which takes roughly
$14$ hours on $4$ L40S GPUs, with the exact wall-clock varying with $\lambda$, model size and
across runs.

\section{Group-Relative Policy Optimization}\label{app:grpo}

For each input $x$, GRPO~\citep{DBLP:journals/corr/abs-2402-03300} samples a group of $G$
candidate outputs $\{o_1, \dots, o_G\}$ from the current policy $\pi_{\theta_{\text{old}}}$
and scores each, giving rewards $\{r_1, \dots, r_G\}$. Writing
$\rho_{i,t}(\theta) = \pi_\theta(o_{i,t}\mid x, o_{i,<t}) /
\pi_{\theta_{\text{old}}}(o_{i,t}\mid x, o_{i,<t})$ for the per-token importance ratio, it
maximizes the clipped surrogate objective
\begin{equation}
  \begin{aligned}
  J(\theta) = \;
  \mathbb{E}_{x,\, \{o_i\} \sim \pi_{\theta_{\text{old}}}}\Bigg[
    \frac{1}{G}\sum_{i=1}^{G} \frac{1}{|o_i|} \sum_{t=1}^{|o_i|} \Big\{
    \min\Big(& \rho_{i,t}(\theta)\, \hat{A}_i, \;
    \operatorname{clip}\!\big(\rho_{i,t}(\theta),\, 1-\epsilon,\, 1+\epsilon\big)\, \hat{A}_i \Big) \\
    &- \kappa \, D_{\text{KL}}\!\left(\pi_\theta \,\|\, \pi_{\text{ref}}\right)
    \Big\} \Bigg],
  \end{aligned}
  \label{eq:grpo-objective}
\end{equation}
where the clipping range $\epsilon$ bounds the size of each update and an optional KL penalty
of weight $\kappa$ keeps the policy close to a reference model $\pi_{\text{ref}}$. The reward
scores a completed sequence, so the advantage is shared by all tokens of a sample and carries
no token index. RS-GRPO~\citep{DBLP:journals/corr/abs-2509-24261} replaces the risk-neutral expected reward with an
exponential utility,
$J_{\text{RS}}(\theta) = \mathbb{E}_{\mathcal{D}}[\,\beta^{-1}\log
\mathbb{E}_{o \sim \pi_\theta}[e^{\beta R(o)}]\,]$, where $\beta \to 0$ recovers the ordinary
expected reward, $\beta > 0$ is risk-seeking and interpolates towards $\max_o R(o)$ as
$\beta \to \infty$, and $\beta < 0$ would be risk-averse. This changes only the advantage, which
becomes
\begin{equation}
  \hat{A}^{\text{RS}}_i \;=\;
  \frac{1}{\beta}\left(
    \frac{e^{\beta r_i}}{\frac{1}{G}\sum_{j=1}^{G} e^{\beta r_j}} - 1
  \right),
  \label{eq:rs-advantage}
\end{equation}
and leaves \eqref{eq:grpo-objective} otherwise unchanged. We use $\beta = 4$ and disable the KL
penalty entirely ($\kappa = 0$), which we found to work best. \appref{app:posttraining-analysis}
shows how the sampled candidates evolve over post-training.

\section{Inference Details}\label{app:inference}

This section adds the details of the inference pipeline of \secref{sec:inference}. When a
candidate has more free constants than data points, which can happen for very small datasets,
the least-squares fit is underdetermined and \texttt{lm} (MINPACK's \texttt{lmdif}) errors, so
we fall back to the trust-region solver \texttt{trf} rather than discard the candidate.
Refinement runs independently per candidate and in parallel, and the post-training reward
reuses it, fitting on a training split before scoring on held-out rows. Restarts perturb the
seed by additive zero-mean Gaussian noise with a standard deviation of $0.5$. Candidates are sampled
at temperature $1.0$ with top-p $0.95$, unchanged in every experiment we report. In general, and
especially in our heavily parameterized setting, constant refinement can be prone to
overfitting. Starting in a predicted constant region reduces that risk without removing it.

\section{Benchmark Datasets}\label{app:datasets}

The twelve black-box datasets of SRBench~2.0 span $47$ to $31{,}104$ rows and $3$ to $51$ columns.
They were curated so as to be neither trivially solvable nor dominated by the Friedman datasets
that were over-represented in the black-box suite of the original SRBench, which makes the track
both harder and more representative of real regression problems. The first-principles track pairs
twelve real or generated problems with an accepted expression.
\tabref{tab:srbench-datasets} lists the datasets of both tracks.

The Feynman and Strogatz tracks are taken unchanged from the original
SRBench~\citep{DBLP:conf/nips/CavaOBFVJKM21}, which we refer to for the individual problems. They
comprise $116$ problems from the Feynman symbolic regression database and $14$ from the ordinary
differential equations of the Strogatz repository, each generated from a known law and evaluated
at four target-noise levels, $0$, $0.001$, $0.01$ and $0.1$. The Feynman problems have
$100{,}000$ rows and up to $9$ features, the Strogatz problems $400$ rows and $2$ features.

\begin{table}[tbp]
  \centering
  \caption{\textbf{The two tracks of SRBench~2.0.} Adapted from the benchmark~\citep{DBLP:conf/gecco/AldeiaZBCFBCF25}. Datasets are listed under the short names used throughout the paper, with the corresponding PMLB identifier given for the black-box track
  and the data source, with the additive noise level for the generated datasets,
  for the first-principles track. The column count includes the target, so a
  dataset with $c$ columns has $c-1$ features. The first-principles identifiers
  are the short names prefixed with \texttt{first\_principles\_}.}
  \label{tab:srbench-datasets}
  \small
  \begin{tabular}{lrrl}
    \hline
    Black-box & Rows & Cols & PMLB identifier \\
    \hline
    SWD           & 1000  & 11 & 1028\_SWD                       \\
    USCrime       & 47    & 14 & 1089\_USCrime                   \\
    lowbwt        & 31104 & 10 & 1193\_BNG\_lowbwt               \\
    echoMonths    & 17496 & 10 & 1199\_BNG\_echoMonths           \\
    vineyard      & 52    & 3  & 192\_vineyard                   \\
    cloud         & 108   & 6  & 210\_cloud                      \\
    pm10          & 500   & 8  & 522\_pm10                       \\
    apnea1        & 475   & 4  & 557\_analcatdata\_apnea1        \\
    fri\_c0\_250\_5  & 250   & 6  & 579\_fri\_c0\_250\_5            \\
    fri\_c2\_1000\_10 & 1000  & 11 & 606\_fri\_c2\_1000\_10          \\
    fri\_c0\_500\_50  & 500   & 51 & 650\_fri\_c0\_500\_50           \\
    environmental & 111   & 4  & 678\_visualizing\_environmental \\
    \hline
    First-principles & Rows & Cols & Source / noise \\
    \hline
    absorption     & 14  & 2 & Russeil et al.~\citep{russeil2024multiview}        \\
    bode           & 8   & 2 & Bonnet~\citep{bonnet1781contemplation}               \\
    hubble         & 32  & 2 & Hubble~\citep{hubble1929relation}               \\
    ideal\_gas     & 30  & 4 & generated, $10\%$ noise           \\
    kepler         & 6   & 2 & Kepler~\citep{Kepler1619}               \\
    leavitt        & 26  & 2 & Leavitt \& Pickering~\citep{leavitt1912periods} \\
    newton         & 30  & 4 & generated, $10\%$ noise           \\
    planck         & 100 & 3 & generated, $10\%$ noise           \\
    rydberg        & 50  & 3 & generated, $1\%$ noise            \\
    schechter      & 27  & 2 & generated, $20\%$ noise           \\
    supernovae\_zr & 236 & 2 & Russeil et al.~\citep{russeil2024multiview}        \\
    tully\_fisher  & 18  & 2 & Tully \& Fisher~\citep{tully1977new}      \\
    \hline
  \end{tabular}
\end{table}

\section{Evaluation Protocol Details}\label{app:protocol}

\paragraph{Runs, tuning and compute.} We follow the evaluation protocol of the respective
benchmark. Every experiment is repeated over a fixed set of distinct seeds, where each seed
determines both the $75/25$ train/test split and any stochasticity in the method. These are the
$30$ seeds of SRBench~2.0 on its black-box and first-principles tracks and the $10$ seeds of the
original SRBench on Feynman and Strogatz. Within a run, hyperparameters are selected by
3-fold cross-validation on the training split. The selected configuration is then refit on the
full training split and evaluated once on the held-out test split. For RASR the only tuned
hyperparameter is the parsimony weight $\gamma$ of \eqref{eq:selection}. We also follow the
benchmark's compute constraint of a single CPU core, here one core of an Intel Xeon Platinum
8562Y+, with the neural methods additionally using one L40S GPU for inference.

\paragraph{Metrics.} The mean test $R^2$ rank depends only on the ordering of methods, so every
dataset contributes equally regardless of how large the gaps on it are. Mean
improvability~\citep{DBLP:journals/corr/abs-2506-16791} is the complementary view, computed from
the RMSE as $(\mathrm{err}_i - \mathrm{err}^\star_i)/\mathrm{err}_i$ per dataset, with
$\mathrm{err}^\star_i$ the lowest error over a fixed pool, so $0\%$ marks the pool-best method and
the magnitude of the gaps is retained. On the first-principles track we score methods by their
per-problem accuracy-complexity Pareto front rather than by a track-level aggregate, and clip test
$R^2$ at $0$ as the benchmark does. The clipping matters for Hubble and Tully-Fisher, where the
accepted expression's own raw test $R^2$ is negative.

\paragraph{Input size handling.} Before fitting we bring each dataset into a range the model can
handle. Rows are capped at $2048$, double the maximum of $1024$ the model was trained on,
subsampling for fitting while still evaluating on the full test split. When subsampling is necessary,
the rows left over form a validation split on which the selection criterion of \eqref{eq:selection} is
computed, replacing training $R^2$. Features are capped at $11$, the maximum the model accepts: we
first drop near-constant features with a variance threshold and, if more than $11$ remain, keep
those most correlated with the target in absolute Pearson correlation. The row cap applies to two
black-box datasets and the feature cap to two further ones, so four of the twelve are reduced in
total. On Feynman, whose problems all come with $100{,}000$ rows, the row cap applies throughout,
while Strogatz and the first-principles track stay within both bounds. Flash-ANSR is subject to the same handling,
with the feature cap raised to the $17$ features it supports; SNIP reduces oversized datasets
internally.

\paragraph{Aggregation.} We deliberately avoid the max-over-seeds aggregation the benchmark uses
for some of its plots, which reports an optimistic best case and rewards high-variance methods and
easier splits rather than typical behavior. We likewise avoid summarizing results through the
area under the performance profile curve that the benchmark reports over both seeds and datasets:
a performance profile is informative precisely because it shows the whole distribution of outcomes
across thresholds, and collapsing it to a single scalar discards what the plot was meant to
convey. \appref{app:perfprofiles} nonetheless reports the black-box track under the benchmark's
own post-processing, so that our runs can be read next to its published numbers. Improvability is
defined relative to a pool of compared methods, and we use two. The Pareto plots against the
baselines use all baselines together with the three RASR variants they show. The ablations and the
remaining studies of our model share a single pool of all baselines together with every RASR
variant appearing in any of them, fixed once rather than recomputed per study, so that their
values are directly comparable.

\paragraph{Baseline configurations.} For the two pre-trained models not part of SRBench~2.0 we
stay as close to the published default configuration as a comparable compute budget allows. For
Flash-ANSR we use the $120$M-parameter model and sample $K = 16{,}384$ candidates, which places it in the upper region of the low
test-time-compute regime its authors propose, and tune the parsimony weight over
$\gamma \in \{0, 0.05, 0.1\}$, which includes their default of $0.05$. For SNIP we keep the
default configuration but limit its latent-space search to a population of $15$ over at most $10$
iterations, again for a comparable budget, and run no hyperparameter search, following the
authors' own SRBench submission.

\section{Handling Failed and Invalid Runs}\label{app:failures}

Not every run returns a usable expression. A run can terminate with an error, exhaust its time
budget without converging, or return an expression that evaluates to \texttt{NaN} on the test
set. In the extreme, every seed of a method can fail on a dataset, as happens for PySR on
USCrime. The benchmark prescribes no consistent handling for these cases. We impute a failed
run conservatively, setting its test $R^2$ to $0$ and its expression size to the median size
over the method's successful runs on that dataset. Where a method fails on every seed, no such
median exists, so the dataset enters the $R^2$ aggregation with a value of $0$ and is left out
of the expression-size aggregation. This is imperfect, but the most neutral rule we found, and
it keeps a method that silently fails from being scored as if it had been absent.
\tabref{tab:app-blackbox-failures} reports how many runs are imputed this way for each method
and dataset on the black-box track. The expression sizes in the released results also contain a small number of
invalid entries recorded as a size of $-1$, which we recompute uniformly with SymPy, matching a
later fix in the SRBench codebase; two individual runs remain unrecoverable and are masked out
of the median aggregation rather than imputed.

\newcommand{\bbhead}[1]{\rotatebox{90}{\footnotesize #1}}
\newcommand{\bbrow}{\\ \hline}
\newcommand{\bbempty}{ & & & & & & & & & & & & \\ \hline}
\colorlet{fclow}{Orange!90!Black}
\colorlet{fchigh}{Red!70!Black}
\newcommand{\fc}[1]{\ifnum#1=0\textcolor{black!35}{0}\else
  \pgfmathtruncatemacro{\fcpct}{round(100*ln(#1)/ln(30))}%
  \textcolor{fchigh!\fcpct!fclow}{\textbf{#1}}\fi}

\begin{table}[p]
  \centering
  \caption{\textbf{Failed runs per black-box dataset.} Number of failed runs out of the $30$
  runs of each method. These are the runs imputed as described above. Zeros are shown in gray, and
  the color of the other counts shifts from orange to dark red with the number of failures.}
  \label{tab:app-blackbox-failures}
  \small
  \setlength{\tabcolsep}{4pt}
  \begin{tabular}{|@{\hspace{3pt}}l|*{12}{r}@{\hspace{3pt}}|}
    \hline
    Method
      & \bbhead{SWD}
      & \bbhead{USCrime}
      & \bbhead{lowbwt}
      & \bbhead{echoMonths}
      & \bbhead{vineyard}
      & \bbhead{cloud}
      & \bbhead{pm10}
      & \bbhead{apnea1}
      & \bbhead{fri\_c0\_250\_5}
      & \bbhead{fri\_c2\_1000\_10}
      & \bbhead{fri\_c0\_500\_50}
      & \bbhead{environmental} \\
    \hline
    RASR-94M-L               & \fc{0} & \fc{0} & \fc{0} & \fc{0} & \fc{0} & \fc{0} & \fc{0} & \fc{0} & \fc{0} & \fc{0} & \fc{0} & \fc{0} \bbrow
    RASR-94M-M               & \fc{0} & \fc{0} & \fc{0} & \fc{0} & \fc{0} & \fc{0} & \fc{0} & \fc{0} & \fc{0} & \fc{0} & \fc{0} & \fc{0} \bbrow
    RASR-94M-S               & \fc{0} & \fc{0} & \fc{0} & \fc{0} & \fc{0} & \fc{0} & \fc{0} & \fc{0} & \fc{0} & \fc{0} & \fc{0} & \fc{0} \bbrow
    AFP                      & \fc{0} & \fc{0} & \fc{0} & \fc{0} & \fc{0} & \fc{0} & \fc{0} & \fc{0} & \fc{0} & \fc{0} & \fc{0} & \fc{0} \bbrow
    AFP\_EHC                 & \fc{0} & \fc{0} & \fc{0} & \fc{0} & \fc{0} & \fc{0} & \fc{0} & \fc{0} & \fc{0} & \fc{0} & \fc{0} & \fc{0} \bbrow
    AFP\_FE                  & \fc{0} & \fc{0} & \fc{0} & \fc{0} & \fc{0} & \fc{0} & \fc{0} & \fc{0} & \fc{0} & \fc{0} & \fc{0} & \fc{0} \bbrow
    Bingo                    & \fc{0} & \fc{0} & \fc{0} & \fc{0} & \fc{0} & \fc{0} & \fc{0} & \fc{0} & \fc{0} & \fc{0} & \fc{0} & \fc{0} \bbrow
    Brush                    & \fc{0} & \fc{0} & \fc{0} & \fc{0} & \fc{0} & \fc{0} & \fc{0} & \fc{0} & \fc{0} & \fc{0} & \fc{0} & \fc{0} \bbrow
    BSR                      & \fc{0} & \fc{0} & \fc{0} & \fc{0} & \fc{0} & \fc{0} & \fc{0} & \fc{0} & \fc{0} & \fc{0} & \fc{0} & \fc{0} \bbrow
    E2E                      & \fc{6} & \fc{6} & \fc{3} & \fc{1} & \fc{0} & \fc{0} & \fc{1} & \fc{2} & \fc{0} & \fc{4} & \fc{1} & \fc{1} \bbrow
    EPLEX                    & \fc{0} & \fc{0} & \fc{0} & \fc{0} & \fc{0} & \fc{0} & \fc{0} & \fc{0} & \fc{0} & \fc{0} & \fc{0} & \fc{0} \bbrow
    EQL                      & \fc{0} & \fc{0} & \fc{0} & \fc{0} & \fc{0} & \fc{0} & \fc{0} & \fc{0} & \fc{0} & \fc{0} & \fc{0} & \fc{0} \bbrow
    FEAT                     & \fc{0} & \fc{0} & \fc{1} & \fc{0} & \fc{0} & \fc{0} & \fc{0} & \fc{0} & \fc{0} & \fc{0} & \fc{0} & \fc{0} \bbrow
    FFX                      & \fc{0} & \fc{0} & \fc{0} & \fc{0} & \fc{0} & \fc{0} & \fc{0} & \fc{0} & \fc{0} & \fc{0} & \fc{0} & \fc{0} \bbrow
    Flash-ANSR               & \fc{0} & \fc{0} & \fc{0} & \fc{1} & \fc{0} & \fc{1} & \fc{0} & \fc{0} & \fc{0} & \fc{0} & \fc{0} & \fc{0} \bbrow
    Genetic Engine           & \fc{0} & \fc{0} & \fc{0} & \fc{0} & \fc{0} & \fc{0} & \fc{0} & \fc{0} & \fc{0} & \fc{0} & \fc{0} & \fc{0} \bbrow
    GP-GOMEA                 & \fc{0} & \fc{0} & \fc{0} & \fc{0} & \fc{0} & \fc{0} & \fc{0} & \fc{0} & \fc{0} & \fc{0} & \fc{0} & \fc{0} \bbrow
    gplearn                  & \fc{0} & \fc{0} & \fc{0} & \fc{0} & \fc{0} & \fc{0} & \fc{0} & \fc{0} & \fc{0} & \fc{0} & \fc{0} & \fc{0} \bbrow
    GPZGD                    & \fc{0} & \fc{0} & \fc{0} & \fc{0} & \fc{0} & \fc{0} & \fc{0} & \fc{0} & \fc{0} & \fc{0} & \fc{0} & \fc{0} \bbrow
    ITEA                     & \fc{0} & \fc{0} & \fc{0} & \fc{0} & \fc{0} & \fc{0} & \fc{0} & \fc{0} & \fc{0} & \fc{0} & \fc{0} & \fc{0} \bbrow
    NeSymReS                 & \fc{0} & \fc{0} & \fc{11} & \fc{11} & \fc{0} & \fc{0} & \fc{0} & \fc{1} & \fc{0} & \fc{0} & \fc{0} & \fc{0} \bbrow
    Operon                   & \fc{0} & \fc{26} & \fc{0} & \fc{0} & \fc{0} & \fc{0} & \fc{0} & \fc{0} & \fc{0} & \fc{0} & \fc{0} & \fc{0} \bbrow
    PS-Tree                  & \fc{0} & \fc{0} & \fc{0} & \fc{0} & \fc{0} & \fc{0} & \fc{0} & \fc{0} & \fc{0} & \fc{0} & \fc{0} & \fc{0} \bbrow
    PySR                     & \fc{0} & \fc{30} & \fc{0} & \fc{0} & \fc{0} & \fc{0} & \fc{0} & \fc{0} & \fc{0} & \fc{0} & \fc{0} & \fc{0} \bbrow
    QLattice                 & \fc{0} & \fc{0} & \fc{0} & \fc{0} & \fc{0} & \fc{0} & \fc{0} & \fc{0} & \fc{0} & \fc{0} & \fc{0} & \fc{0} \bbrow
    Rils-Rols                & \fc{0} & \fc{0} & \fc{0} & \fc{0} & \fc{0} & \fc{0} & \fc{0} & \fc{0} & \fc{0} & \fc{0} & \fc{0} & \fc{0} \bbrow
    SNIP                     & \fc{0} & \fc{1} & \fc{0} & \fc{0} & \fc{0} & \fc{0} & \fc{0} & \fc{0} & \fc{0} & \fc{0} & \fc{0} & \fc{0} \bbrow
    TIR                      & \fc{0} & \fc{0} & \fc{0} & \fc{0} & \fc{0} & \fc{0} & \fc{0} & \fc{0} & \fc{0} & \fc{0} & \fc{0} & \fc{0} \bbrow
    TPSR                     & \fc{0} & \fc{0} & \fc{11} & \fc{11} & \fc{0} & \fc{0} & \fc{0} & \fc{0} & \fc{0} & \fc{0} & \fc{0} & \fc{0} \bbrow
    uDSR                     & \fc{0} & \fc{0} & \fc{0} & \fc{0} & \fc{0} & \fc{0} & \fc{0} & \fc{0} & \fc{0} & \fc{0} & \fc{0} & \fc{0} \bbrow
  \end{tabular}
\end{table}

\section{Additional Black-Box Results}\label{app:additional-blackbox}

\begin{figure}[!ht]
  \centering
  \includegraphics[width=\linewidth]{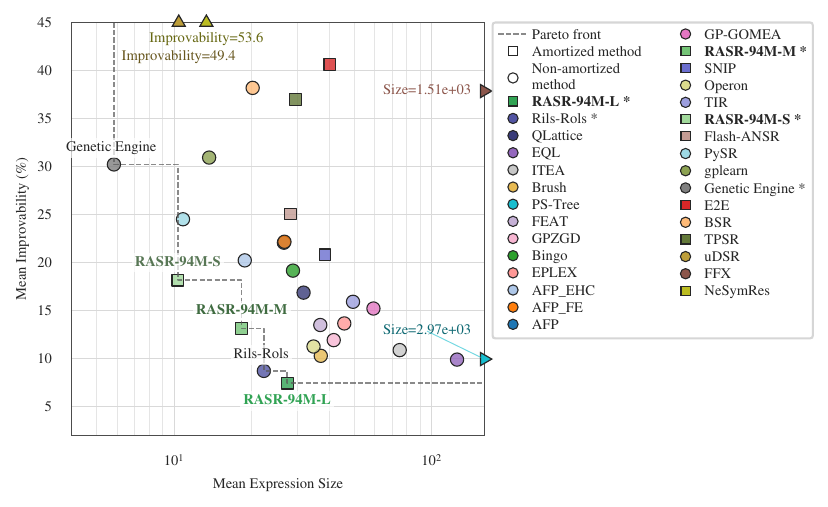}
  \caption{\textbf{Accuracy-complexity trade-off with all baselines.} As
  \figref{fig:blackbox-pareto}, with every evaluated baseline shown individually
  instead of only the methods named on the plot.}
  \label{fig:app-pareto-full}
\end{figure}

\paragraph{Full per-method Pareto front.} \figref{fig:app-pareto-full} gives the same
accuracy-complexity trade-off as \figref{fig:blackbox-pareto}, but with every evaluated baseline
identified individually rather than only the handful named directly on the main figure.

\paragraph{Additional variants.} The figures below complement \secref{subsec:blackbox}, adding
RASR-24M-L, RASR-94M-base, RASR-94M-M, and RASR-94M-S and, in two of the
three, restricting the track to the eight datasets within our input bounds (\secref{subsec:setup}).
Throughout, each RASR variant is ranked against the baselines alone, independently of the other
variants, so that the variants do not displace one another. RASR-24M-L is second
over all twelve datasets, which is the result the ablations at that scale rest on.

\begin{figure}[tbp]
  \centering
  \includegraphics[width=\linewidth]{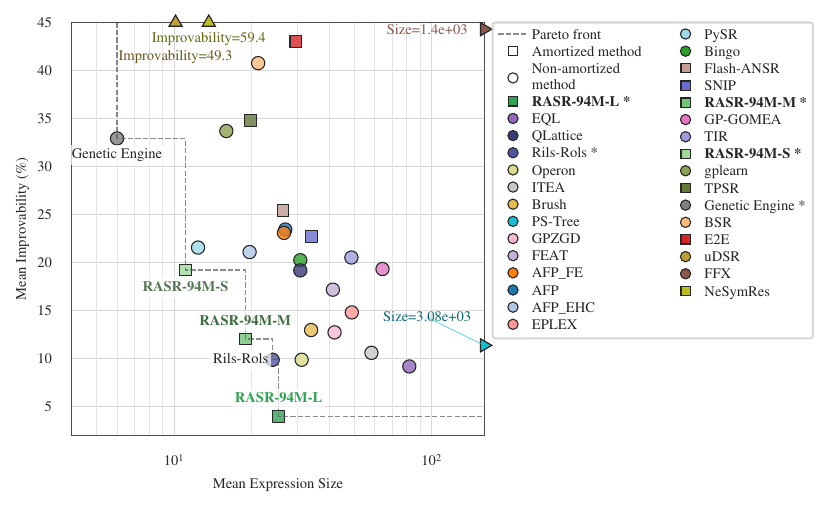}
  \caption{\textbf{Accuracy-complexity trade-off on unreduced datasets.} As
  \figref{fig:blackbox-pareto}, restricted to the eight datasets
  RASR neither subsamples nor feature-reduces.}
  \label{fig:app-pareto-small}
\end{figure}

\paragraph{Datasets within our input bounds.} On the eight datasets RASR neither subsamples nor
feature-reduces, both aggregate views put it further ahead. RASR-94M-L and RASR-24M-L take the first
two mean-rank positions with a visible gap to EQL behind them
(\figref{fig:app-overview-small}), and on the Pareto plot RASR-94M-L drops to roughly $4\%$ mean
improvability, against $7.4\%$ over all twelve, while the next baseline, Rils-Rols, stays just
under $10\%$ (\figref{fig:app-pareto-small}). The size-penalized variants move with it, to
roughly $12\%$ at $19$ nodes and $19\%$ at $11$. This is the regime the model was trained for, so
the four reduced datasets cost accuracy.

\begin{figure}[tbp]
  \centering
  \includegraphics[width=\linewidth]{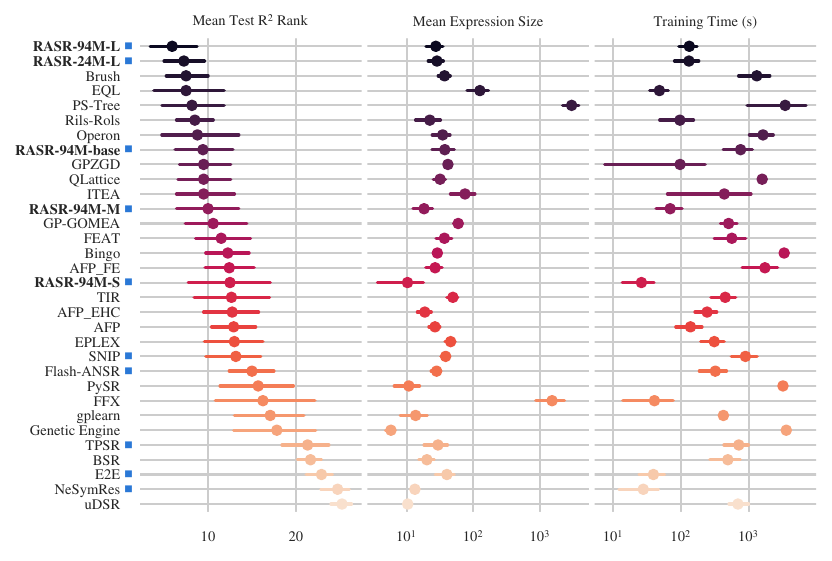}
  \caption{\textbf{Black-box track overview with all baselines and RASR variants.} As
  \figref{fig:blackbox-overview} over all twelve datasets, but with every baseline shown and with
  RASR-24M-L, RASR-94M-base, RASR-94M-M, and RASR-94M-S alongside RASR-94M-L. Each RASR variant is ranked against
  the baselines alone, so the baseline ranks match \figref{fig:blackbox-overview}.}
  \label{fig:app-overview-variants}
\end{figure}

\begin{figure}[tbp]
  \centering
  \includegraphics[width=\linewidth]{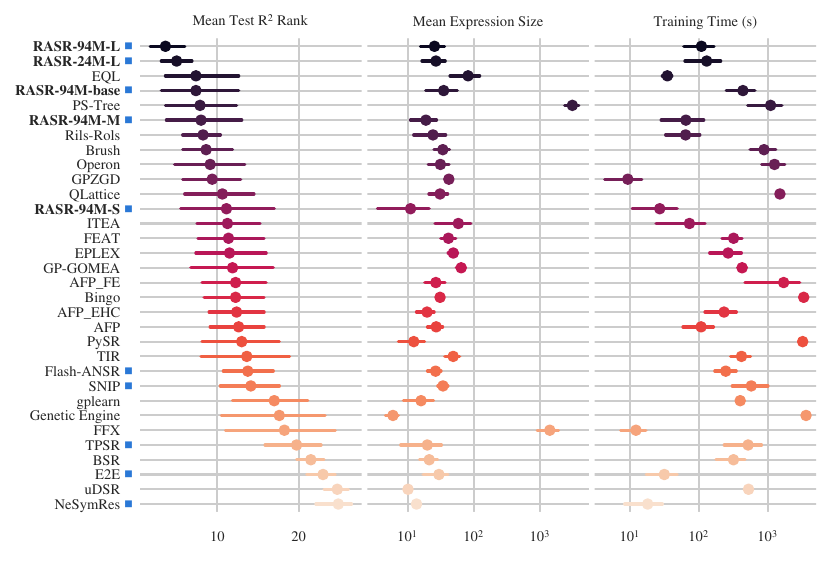}
  \caption{\textbf{Black-box track overview on unreduced datasets.} As
  \figref{fig:app-overview-variants}, restricted to the eight datasets
  RASR neither subsamples nor feature-reduces.}
  \label{fig:app-overview-small}
\end{figure}

\section{Additional Ground-Truth Results}\label{app:groundtruth}

\paragraph{Behavior under noise.} \secref{subsec:feynman} reports that noise is what separates the
methods on these tracks. The full per-noise numbers behind that statement are as follows. On
Strogatz, RASR-94M-L solves $99.3\%$ of the fourteen problems without noise and still $65.0\%$ at
$10\%$ target noise, against $41.4\%$ for Flash-ANSR and $32.1\%$ for GP-GOMEA at that level, while
Operon falls from $86.4\%$ to $5.0\%$ and SBP-GP and MRGP to zero. On Feynman, RASR degrades from
$78.9\%$ to $72.5\%$ over the same range, where Flash-ANSR drops from $89.5\%$ to $70.5\%$ and
AIFeynman from $73.2\%$ to $9.9\%$. Expression size shows the same pattern. RASR returns $34$ nodes
on Feynman at every noise level, whereas Operon grows from $70$ to $89$ nodes and MRGP from $3.2$k
to $86$k as they fit the added noise.

\paragraph{Feynman problems by number of variables.} Of the $116$ Feynman problems, $102$ have at
most five variables in the ground-truth law and $14$ have six or more. Without noise RASR-94M-L
solves $86.7\%$ of the problems with at most five variables, against $92.2\%$ for Flash-ANSR and
$90.4\%$ for Operon, but only $22.1\%$ of those with six or more, against $70.0\%$ and $47.1\%$.
Averaged over the four noise levels, RASR-94M-L comes close to both on the problems with at most
five variables, at $85.1\%$ against $87.8\%$ for each, while on those with six or more the gap
remains, at $19.3\%$ against $54.1\%$ and $46.8\%$. Of the Feynman runs RASR-94M-L
does not solve, $70\%$ still reach $R^2 > 0.99$.

\begin{figure}[htbp]
  \centering
  \begin{minipage}[t]{0.498\linewidth}
    \centering
    \includegraphics[width=\linewidth]{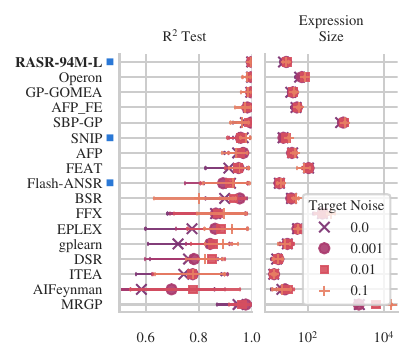}
  \end{minipage}\hfill
  \begin{minipage}[t]{0.498\linewidth}
    \centering
    \includegraphics[width=\linewidth]{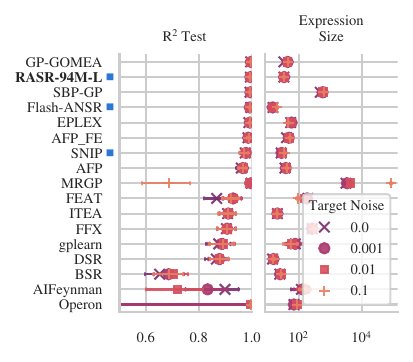}
  \end{minipage}
  \caption{\textbf{\boldmath Test $R^2$ on the ground-truth tracks.} As \figref{fig:groundtruth} for the
  Strogatz (left) and Feynman (right) tracks, but
  scoring mean test $R^2$ (higher is better) rather than the accuracy solution rate. The $R^2$
  axis is clipped at $0.5$, so a method with a lower mean at some noise level falls outside it.
  Amortized methods are flagged with a blue square next to their name.}
  \label{fig:app-groundtruth-r2}
\end{figure}

\paragraph{Test $R^2$ view.} \figref{fig:app-groundtruth-r2} shows the same runs under median test
$R^2$ instead of the accuracy solution rate, which also credits the problems a method gets close on
without solving outright. The two views agree at the top but compress the gaps. On Strogatz
RASR-94M-L is again first, at a mean test $R^2$ of $0.999$ against $0.996$ for Operon and $0.994$ for
GP-GOMEA, and on Feynman it is level with SBP-GP at $0.995$, behind GP-GOMEA's $0.997$ and ahead of
Flash-ANSR. The methods that collapse under $10\%$ target noise in the solution-rate view collapse
here as well, AIFeynman and MRGP to negative mean $R^2$ on both tracks.

\section{First-Principles Results}\label{app:firstprinciples}

\begin{figure}[htbp]
  \centering
  \includegraphics[width=\linewidth]{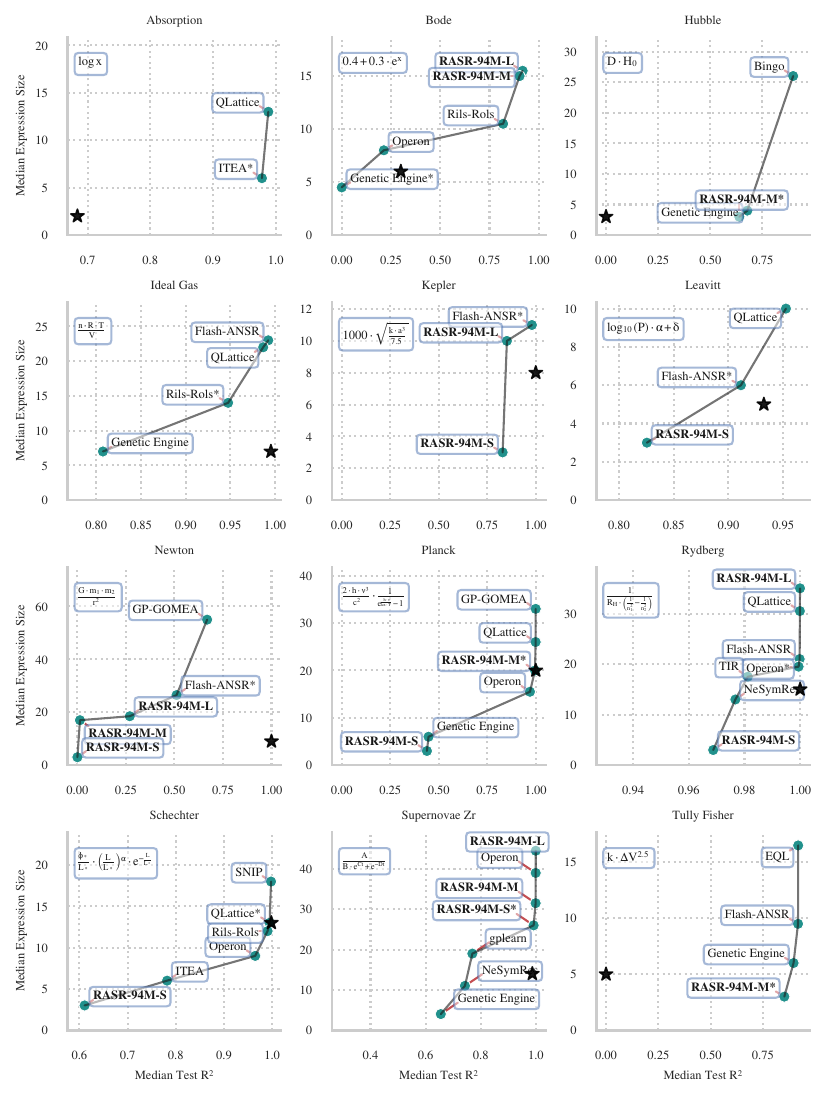}
  \caption{\textbf{First-principles track.} Per-problem accuracy-complexity Pareto fronts of median
  expression size against median test $R^2$. Only non-dominated methods appear. The black star is the accepted expression.}
  \label{fig:firstprinciples-pareto}
\end{figure}

The first-principles track of SRBench~2.0 pairs each of its twelve problems with an accepted
expression, as the ground-truth tracks do, but its data are real historical measurements or
generated with up to $20\%$ noise, and the problems are tiny, from six to $236$ rows. The accepted
expression is therefore not always the best predictor, which lets us ask how a method's
expressions relate to it rather than only how well they fit. \figref{fig:firstprinciples-pareto}
gives the per-problem Pareto fronts of median expression size against median test $R^2$, with the
accepted expression marked.

\paragraph{Front coverage.} At least one RASR variant is on the front for ten of the twelve
problems, the exceptions being Absorption and Ideal Gas. Counted per variant,
RASR-94M-S appears on seven fronts, with RASR-94M-M, Flash-ANSR, QLattice
and Genetic Engine next at six and RASR-94M-L at five.

\paragraph{Fit at the law's size.} On the generated problems the data follows the law up to
additive noise, so the law is close to the best attainable $R^2$ and what separates methods is the
size at which they reach it. The three variants cover different parts of that range. On Rydberg,
RASR-94M-S reaches $R^2 = 0.97$ with three nodes against the law's fifteen, while the
default RASR-94M-L sits at the other end of the same front at $R^2 = 0.9999$ and $35$ nodes. On
Planck, RASR-94M-M reaches $R^2 = 0.996$ at exactly the accepted expression's size of $20$ nodes,
where QLattice and GP-GOMEA need $26$ and $33$.

\section{Post-Training Analysis}\label{app:posttraining-analysis}

All improvability numbers in this and the next section are on the black-box track (\%, lower is
better), computed against the shared pool of all baselines and every RASR variant
(\appref{app:protocol}).

\begin{figure}[htbp]
  \centering
  \includegraphics[width=\linewidth]{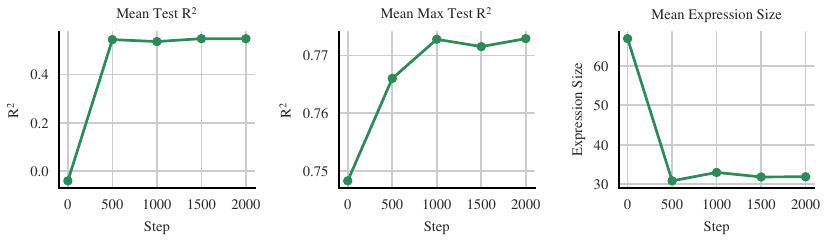}
  \caption{\textbf{Post-training dynamics of RASR-94M-L.} Mean $R^2$ of the sampled candidates,
  mean best-of-$256$ $R^2$ and mean expression size over post-training steps, on a held-out
  validation set drawn from the post-training prior.}
  \label{fig:posttraining-curves}
\end{figure}

\paragraph{Training dynamics.} \figref{fig:posttraining-curves} tracks the sampled candidates of
RASR-94M-L over post-training on a held-out validation set drawn from the post-training prior,
evaluated every $500$ steps. Most of the change happens before the first evaluation. The mean $R^2$
of individual candidates rises from slightly below zero to about $0.55$, and the mean expression
size halves from about $67$ to about $32$ nodes, after which both stay flat. The best-of-$256$
$R^2$ moves far less, from $0.75$ to $0.77$, and keeps rising slightly until step $1000$.

\paragraph{Accuracy, expression size and fitting time.} Post-training changes all three quantities
of \figref{fig:blackbox-overview}. \tabref{tab:posttraining-size-time} gives expression size and
fitting time, aggregated in the same way, as the median over seeds and the mean over datasets. At
the $94$M scale post-training lowers improvability from $10.03$ to $7.44$, shortens the returned
expressions from $37.4$ to $27.5$ nodes, by $27\%$, and cuts the fitting time from $762$ to $133$
seconds, by $83\%$. The larger size penalties of RASR-94M-M and RASR-94M-S push size and time
further, to $18$ and $10$ nodes and $69$ and $26$ seconds, at the cost of accuracy. At $24$M the
size and time reductions are nearly the same, from $39.3$ to $28.5$ nodes and from $646$ to $131$
seconds, while improvability stays level at $9.36$ against $9.51$. The time saving follows the
shorter expressions, which are cheaper to refine, and presumably also the sharper policy proposing
fewer distinct candidates, since duplicates are refined only once (\secref{sec:inference}).

\begin{table}[htbp]
  \caption{\textbf{Effect of post-training on expression size and fitting time on the black-box
  track.} Mean expression size (nodes) and fitting time (seconds), both lower is better, for the
  pre-trained-only base model and the post-trained model at $K = 1024$ sampled candidates.}
  \label{tab:posttraining-size-time}
  \begin{center}
  \begin{tabular}{lcccc}
    \toprule
    & \multicolumn{2}{c}{\bf Expression size} & \multicolumn{2}{c}{\bf Fitting time (s)} \\
    \cmidrule(lr){2-3} \cmidrule(lr){4-5}
    {\bf Variant} & {\bf Base} & {\bf Post-trained} & {\bf Base} & {\bf Post-trained} \\
    \midrule
    RASR-94M-L & 37.4 & 27.5 & 762 & 133 \\
    RASR-94M-M & 37.4 & 18.3 & 762 & 69 \\
    RASR-94M-S & 37.4 & 10.3 & 762 & 26 \\
    RASR-24M-L & 39.3 & 28.5 & 646 & 131 \\
    \bottomrule
  \end{tabular}
  \end{center}
\end{table}

\begin{figure}[tbp]
  \centering
  \includegraphics[width=0.6\linewidth]{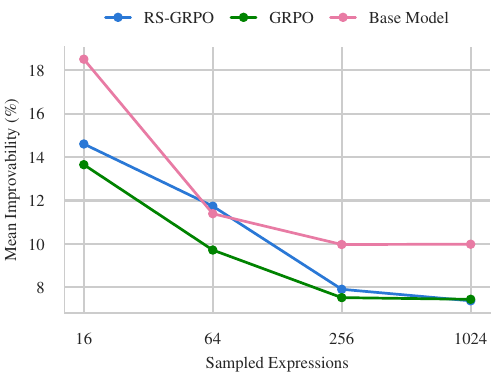}
  \caption{\textbf{Accuracy versus test-time compute on the black-box track.} Mean improvability (\%,
  lower is better) against the number of sampled candidates $K$ for the base model, a model
  post-trained with GRPO, and the main model post-trained with RS-GRPO, all at the $94$M scale.}
  \label{fig:posttraining-sweep}
\end{figure}

\paragraph{Test-time compute.} \figref{fig:posttraining-sweep} plots mean improvability against
the number of sampled candidates $K$ for the pre-trained-only base model, a model post-trained
with standard GRPO, and the main model post-trained with RS-GRPO. We vary $K$ while keeping the other hyperparameters fixed,
so that the plot isolates the
effect of the sample budget. All three models improve with more samples. Apart from RS-GRPO at
$K = 64$,
post-training improves on the base model at every budget. The two post-training objectives end
up close. RS-GRPO is ahead of GRPO on most individual datasets at the full budget
of $1024$ candidates, but a small number of datasets carry the aggregate the other way, USCrime in
particular, a small and noisy dataset that is also one of the four we have to reduce to fit our
input bounds. We
therefore do not claim a separation between the two objectives at large budgets from these runs.

\section{Ablations}\label{app:ablations}

The ablations require pre-training variants of the model, so they run at the $24$M scale, where
the lower cost makes them affordable. Improvability is computed as in
\appref{app:posttraining-analysis}.

\paragraph{Synthetic prior.} This ablation isolates the contribution of our synthetic prior. We
compare it against a simple prior in the style of earlier neural symbolic-regression work, with
inputs drawn from independent uniform distributions, constants sampled from fixed ranges, and no
standardization, and hold the rest of the pipeline fixed. \tabref{tab:prior-ablation} gives the
$2\times2$ grid over the prior and post-training discussed in \secref{subsec:further}.
Post-training data for the simple-prior variant is drawn from the simple prior as well, so that it
is a fair, fully self-contained baseline. Standardization is bundled into our prior here rather
than ablated on its own axis, since it is a defining part of the recipe and only makes sense
alongside the standardized inputs. Our prior takes the base model from $26.56$ to $9.36$.
Post-training helps the simple prior more, lowering it to $18.31$, but leaves it at about twice
the improvability of our prior, which already beats it without any post-training.

\begin{table}[htbp]
  \caption{\textbf{\boldmath Prior ablation on the black-box track ($24$M scale).} The simple prior draws inputs from
  independent uniform distributions and constants from fixed ranges, and applies no
  standardization. Mean improvability (\%, lower is better).}
  \label{tab:prior-ablation}
  \begin{center}
  \begin{tabular}{lcc}
    \toprule
    {\bf Prior} & {\bf Base} & {\bf Post-trained} \\
    \midrule
    Simple prior & 26.56 & 18.31 \\
    Our prior    & 9.36  & 9.51 \\
    \bottomrule
  \end{tabular}
  \end{center}
\end{table}

\paragraph{Constant strategy.} This ablation validates the constants-as-initialization strategy
of \secref{sec:prior}, which places RASR between skeleton-only prediction and full end-to-end
constant prediction. We compare against three alternatives: predicting constants but trusting
them directly, without the least-squares refinement; refining them without initialization
noise; and a skeleton-only variant, trained with our pipeline
but predicting placeholder constants. Refinement is by far the most important component:
removing it raises improvability from $9.51$ to $28.33$, undoing most of the model's advantage.
Dropping the initialization noise raises it to $10.57$. The skeleton-only variant reaches $12.09$, behind both, so predicting constants as
an initialization is worth a real gain over refining placeholder constants from scratch.

\begin{table}[tbp]
  \centering
  \caption{\textbf{\boldmath Constant-strategy ablation on the black-box track ($24$M scale).} Mean
  improvability (\%, lower is better) for our constants-as-initialization approach
  against a skeleton-only variant, refinement without initialization noise, and no
  least-squares refinement. The last two rows are RASR-24M-L with the respective inference step
  disabled. The skeleton-only variant is pre-trained separately and post-trained with the same
  size penalty.}
  \label{tab:constants-ablation}
  \small
  \begin{tabular}{lc}
    \hline
    Variant & Improvability \\
    \hline
    RASR-24M-L                   & 9.51 \\
    Skeleton-only                & 12.09 \\
    No init noise  & 10.57 \\
    No refinement                & 28.33 \\
    \hline
  \end{tabular}
\end{table}

\section{SRBench~2.0 Performance Profiles}\label{app:perfprofiles}

The benchmark reports the black-box track through performance profiles, curves giving the
fraction of problems on which a method reaches at least a given test $R^2$, summarized by the area
under them. We use mean rank and mean improvability instead, for the reasons given in
\appref{app:protocol}. This section adds the benchmark's own view of our runs, produced with its
post-processing code and its metric, so that the numbers here can be read next to the published
SRBench~2.0 results without any translation.

\begin{figure}[tbp]
  \centering
  \includegraphics[angle=90]{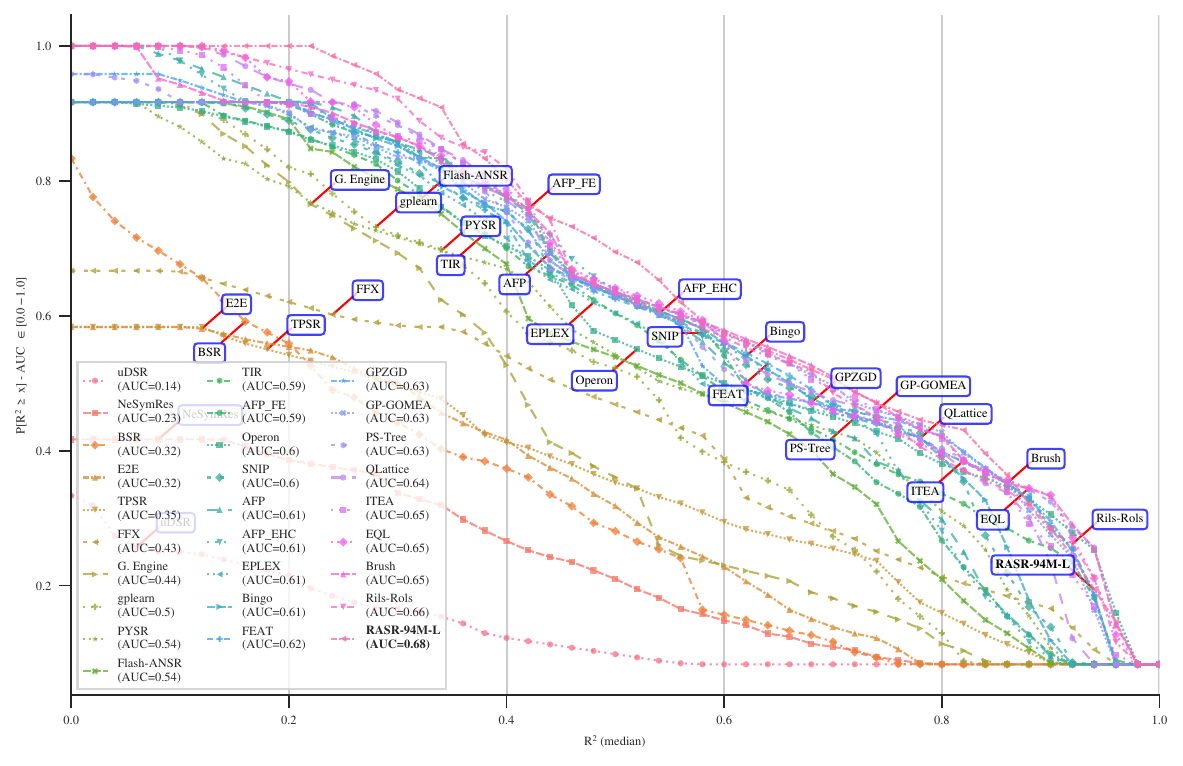}
  \caption{\textbf{Performance profile on the black-box track.} The fraction of problems on which
  a method reaches at least a given test $R^2$, computed per seed and
  aggregated by the median of the $30$ per-seed curves at every threshold, using the SRBench~2.0
  post-processing. The legend gives the area under each
  curve. Higher is better throughout.}
  \label{fig:app-perfprof-median}
\end{figure}

\begin{figure}[tbp]
  \centering
  \includegraphics[width=\linewidth]{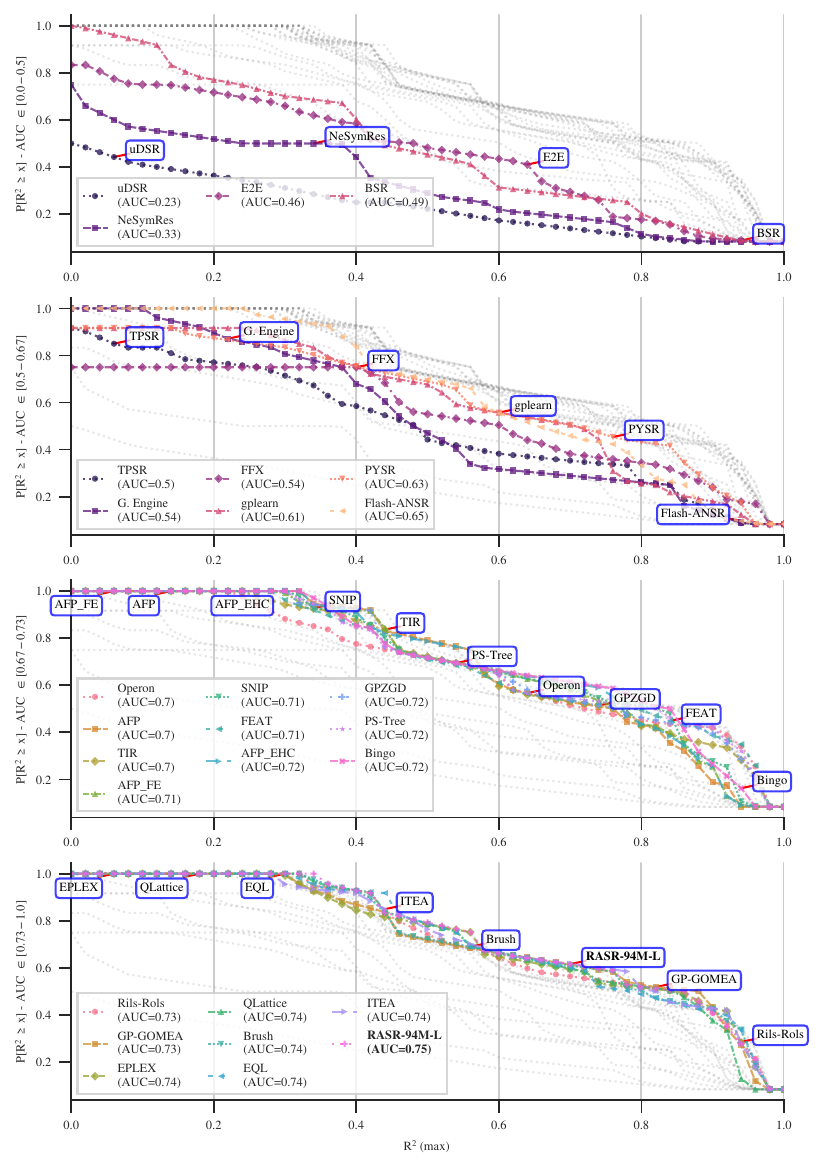}
  \caption{\textbf{Performance profile with the maximum over seeds.} As
  \figref{fig:app-perfprof-median}, but taking the maximum instead of the median of the per-seed
  curves, the aggregation the benchmark uses. Methods are split into four panels by their area under the
  curve so that the curves stay legible, with the methods outside a panel's band drawn in gray for
  reference.}
  \label{fig:app-perfprof-intervals}
\end{figure}

\figref{fig:app-perfprof-median} computes one profile per seed over the twelve datasets and takes,
at every $R^2$ threshold, the median of these curves over the $30$ seeds. RASR-94M-L has the highest area of all evaluated methods at $0.68$,
ahead of Rils-Rols at $0.66$ and a group of four at $0.65$, and its curve is at or above every
other one across most of the $R^2$ range. \figref{fig:app-perfprof-intervals} instead takes the
maximum of the per-seed curves and splits the field into four
bands by area so the curves stay readable. RASR-94M-L again has the highest area, $0.75$ against
$0.74$ for GP-GOMEA, EQL, QLattice, Brush, ITEA and EPLEX. The lead is narrower under this
aggregation, which is the expected effect of scoring every method by its best seed at every
threshold.

\end{document}